%% file: main.tex
\documentclass[10pt,twocolumn,letterpaper]{article}
\usepackage[pagenumbers]{cvpr} 

\input{tex/plots}

\usepackage{colortbl}
\usepackage{xcolor}
\usepackage{multirow}
\usepackage[export]{adjustbox}
\usepackage{xhfill}
\usepackage[T1]{fontenc}

\input{tex/abbrev}
\input{tex/macro}

\usepackage{pifont}
\newcommand{\cmark}{\ding{51}}%
\newcommand{\xmark}{\ding{55}}%

\newcommand{\parag}[1]{\smallskip\noindent\textbf{#1}\enspace}

\newcommand{\myig}[1]{\includegraphics[width=0.118\textwidth,valign=c]{#1}}

\newcommand{\titletext}{{SPOT: \underline{S}elf-Training with \underline{P}atch-Order Permutation for\\ \underline{O}bject-Centric Learning with Autoregressive \underline{T}ransformers}}

\definecolor{cvprblue}{rgb}{0.21,0.49,0.74}
\usepackage[pagebackref,breaklinks,colorlinks,citecolor=cvprblue]{hyperref}

\title{\titletext}

\author{Ioannis~Kakogeorgiou$^{1}$ \hspace{0.5em} Spyros~Gidaris$^{2}$ \hspace{0.5em} Konstantinos~Karantzalos$^{1}$ \hspace{0.5em} Nikos~Komodakis$^{3,4,5}$ \vspace{0.5em} 
\\
$^1$National Technical University of Athens \hspace{1.0em} $^2$valeo.ai \\
$^3$University of Crete \hspace{1.0em} $^4$IACM-Forth \hspace{1.0em} $^5$Archimedes/Athena RC
}

\begin{document}

\input{tex/teaser}

\maketitle

\input{tex/0_abstract}    
\input{tex/1_intro}
\input{tex/2_related}
\input{tex/3_method}
\input{tex/4_experiments}
\input{tex/5_conclusion}
\input{tex/acknowledgements}

{
\small
\bibliographystyle{ieeenat_fullname}
\bibliography{main}
}

\clearpage

\appendix
\input{tex/supplementary}

\end{document}

%% file: tex/plots.tex
\usepackage[dvipsnames,svgnames,x11names]{xcolor}
\usepackage{tikz}
\usetikzlibrary{arrows.meta,shapes,calc,matrix,fit,positioning,backgrounds,decorations.markings}
\usepackage{pgfplots}
\usepackage{pgfplotstable}
\pgfplotsset{compat=1.9}
\usepackage{xstring}

\usepgfplotslibrary{external}
\newcommand{\extfig}[2]{\tikzsetnextfilename{#1}{#2}}

\IfBeginWith*{\jobname}{fig/extern/}{\finalcopy}{}

\tikzstyle{every picture}+=[
	remember picture,
	every text node part/.style={align=center},
	every matrix/.append style={ampersand replacement=\&},
]
\tikzstyle{tight} = [inner sep=0pt,outer sep=0pt]
\tikzstyle{node}  = [draw,circle,tight,minimum size=12pt,anchor=center]
\tikzstyle{op}    = [draw,circle,tight]
\tikzstyle{dot}   = [fill,draw,circle,inner sep=1pt,outer sep=0]
\tikzstyle{pt}    = [fill,draw,circle,inner sep=1.5pt,outer sep=.2pt]
\tikzstyle{box}   = [draw,rectangle,inner sep=3pt]
\tikzstyle{high}  = [black!60]
\tikzstyle{group} = [high,box,opacity=.5]
\tikzstyle{dim1}  = [fill opacity=.3,text opacity=1]
\tikzstyle{dim2}  = [fill opacity=.5,text opacity=1]
\tikzstyle{dim3}  = [fill opacity=.7,text opacity=1]
\tikzstyle{rectc} = [tight,transform shape]
\tikzstyle{rect}  = [rectc,anchor=south west]

\tikzset{every mark/.append style={solid}}
\pgfplotsset{
	grid=both, width=\columnwidth, try min ticks=5,
	every axis/.append style={font=\small},
	every axis plot/.append style={thick,mark=none,mark size=1.8,tension=0.18},
	legend cell align=left, legend style={fill opacity=0.8},
	xticklabel={\pgfmathprintnumber[assume math mode=true]{\tick}},
	yticklabel={\pgfmathprintnumber[assume math mode=true]{\tick}},
	nodes near coords math/.style={
		nodes near coords={\pgfmathprintnumber[assume math mode=true]{\pgfplotspointmeta}},
	},
}

\pgfplotsset{
	dash/.style={mark=o,dashed,opacity=0.6},
	dott/.style={mark=o,dotted,opacity=0.6},
	nolim/.style={enlargelimits=false},
	plain/.style={every axis plot/.append style={},nolim,grid=none},
}

\pgfdeclarelayer{bg4}
\pgfdeclarelayer{bg3}
\pgfdeclarelayer{bg2}
\pgfdeclarelayer{bg1}
\pgfdeclarelayer{fg1}
\pgfdeclarelayer{fg2}
\pgfdeclarelayer{fg3}
\pgfdeclarelayer{fg4}
\pgfsetlayers{bg4,bg3,bg2,bg1,main,fg1,fg2,fg3,fg4}

\pgfkeys{/tikz/.cd, aspect/.store in=\aspect, aspect=1}
\pgfkeys{/tikz/.cd, depth/.store in=\depth, depth=.5}
\pgfkeys{/tikz/.cd, stepx/.store in=\step, stepx=1}

\tikzstyle{geom} = [line join=bevel,aspect=1,depth=.5,z={(\depth*\aspect,\depth)}]
\tikzstyle{wire} = [geom,draw,thick]

\def\cx[#1,#2,#3]{#1}
\def\cy[#1,#2,#3]{#2}
\def\cz[#1,#2,#3]{#3}
\def\ex[#1,#2,#3]{#1,0,0}
\def\ey[#1,#2,#3]{0,#2,0}
\def\ez[#1,#2,#3]{0,0,#3}



%% file: tex/abbrev.tex
\newcommand{\softmax}{{\textsc{Softmax}}}

\newcommand{\transformerdecoder}{{\textsc{Decoder}}}
\newcommand{\bos}{{\textsc{[BOS]}}}

\newcommand{\Th}[1]{\textsc{#1}}
\newcommand{\mr}[2]{\multirow{#1}{*}{#2}}
\newcommand{\mc}[2]{\multicolumn{#1}{c}{#2}}

\newcommand{\red}[1]{{\textcolor{red}{#1}}}

\newcommand{\citeme}[1]{\red{[XX]}}
\newcommand{\refme}[1]{\red{(XX)}}

\newcommand{\fig}[2][1]{\includegraphics[width=#1\linewidth]{fig/#2}}

\newcommand{\tran}{^\top}

\newcommand{\real}{\mathbb{R}}

\newcommand{\vu}{\mathbf{u}}

\newcommand{\vy}{\mathbf{y}}

\newcommand{\mbo}{\Th{mBO}\xspace}%
\newcommand{\mboi}{\Th{mBO$^{i}$}\xspace}%
\newcommand{\mboc}{\Th{mBO$^{c}$}\xspace}%
\newcommand{\fgari}{\Th{FG-ARI}\xspace}%
\newcommand{\fgiou}{\Th{mIoU}\xspace}%
\newcommand{\miou}{\Th{mIoU}\xspace}%

\newcommand{\decoder}{\Th{Decoder}\xspace}
\newcommand{\encoder}{\Th{Slot Attention}\xspace}
\newcommand{\maxdecenc}{\Th{Max$($Dec, Slot Att$)$}\xspace}

\makeatletter
\newcommand*\bdot{\mathpalette\bdot@{.7}}
\newcommand*\bdot@[2]{\mathbin{\vcenter{\hbox{\scalebox{#2}{$\m@th#1\bullet$}}}}}
\makeatother

\makeatletter
\DeclareRobustCommand\onedot{\futurelet\@let@token\@onedot}
\def\@onedot{\ifx\@let@token.\else.\null\fi\xspace}

\def\ie{\emph{i.e}\onedot}

\makeatother

%% file: tex/macro.tex
\newcommand{\Note}[2]{}

\definecolor{ForestGreen}{RGB}{34,139,34}
\definecolor{Orange}{RGB}{1.0, 0.55, 0.0}
\definecolor{Purple}{RGB}{0.58, 0.44, 0.86}

\definecolor{TableColor}{rgb}{0.835, 0.894, 0.968}

%% file: tex/teaser.tex
\makeatletter
\apptocmd\@maketitle{{\teaser{}}}{}{}
\makeatother

\newcommand{\teaser}{%
\vspace{-6pt}
\centering
\setlength{\tabcolsep}{1pt}
\begin{tabular}{@{}ccccccccc@{}}
    \myig{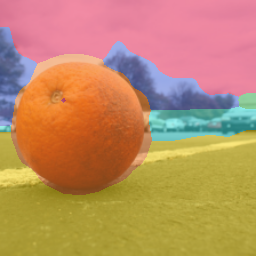} & \myig{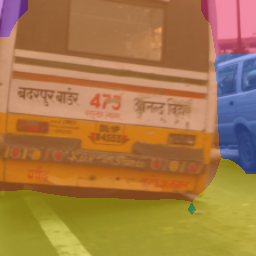} & \myig{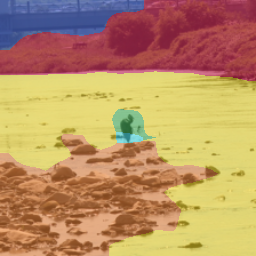} & 
    \myig{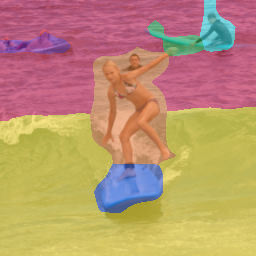} & \myig{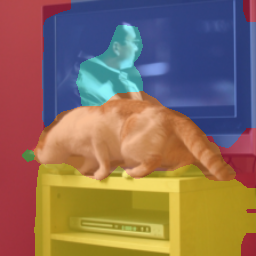} & \myig{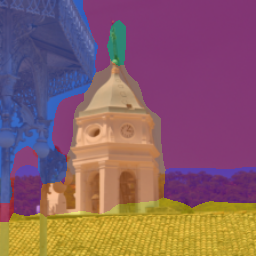} &
    \myig{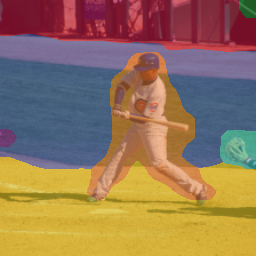} & \myig{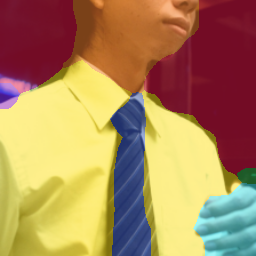} \\
    
    \addlinespace[2.5pt] 
    \myig{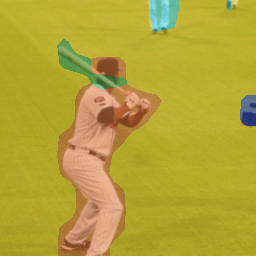} & \myig{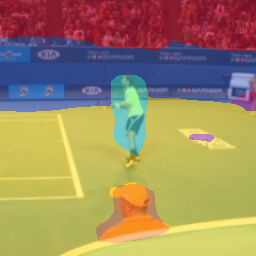} & \myig{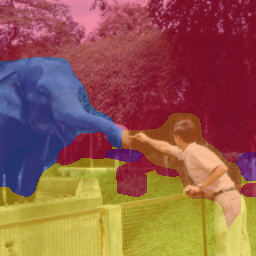} &
    \myig{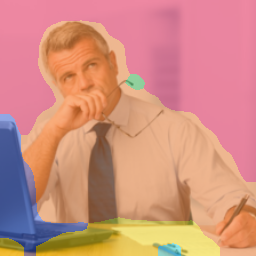} & \myig{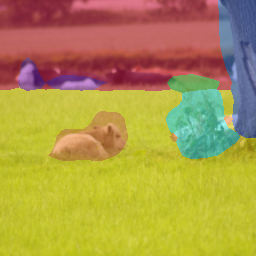} & \myig{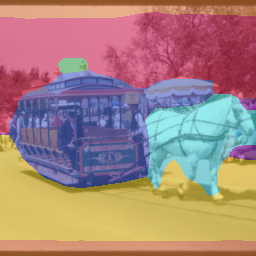} & \myig{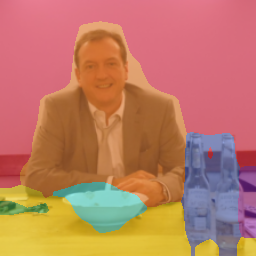} & 
    \myig{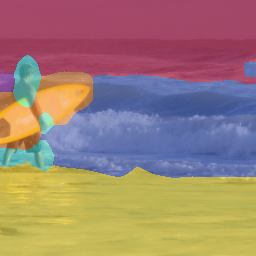} \\

\end{tabular}
\vspace{-6pt}
\captionof{figure}{SPOT: Our novel framework enhances unsupervised object-centric learning in slot-based autoencoders using self-training and sequence permutations in the transformer decoder. It improves object-specific slot generation, excelling in complex real-world images.}
\label{fig:teaser}
\par\vspace{24pt}
}

%% file: tex/0_abstract.tex
\begin{abstract}
Unsupervised object-centric learning aims to decompose scenes into interpretable object entities, termed slots. Slot-based auto-encoders stand out as a prominent method for this task. Within them, crucial aspects include guiding the encoder to generate object-specific slots and ensuring the decoder utilizes them during reconstruction. This work introduces two novel techniques, (i) an attention-based self-training approach,  which distills superior slot-based attention masks from the decoder to the encoder, enhancing object segmentation, and (ii) an innovative patch-order permutation strategy for autoregressive transformers that strengthens the role of slot vectors in reconstruction. The effectiveness of these strategies is showcased experimentally. The combined approach significantly surpasses prior slot-based autoencoder methods in unsupervised object segmentation, especially with complex real-world images. We provide the implementation code at \href{https://github.com/gkakogeorgiou/spot}{https://github.com/gkakogeorgiou/spot}.
\end{abstract}

%% file: tex/1_intro.tex
\section{Introduction}
\label{sec:intro}

Decomposing a scene into separate objects is crucial for AI progress. While current AI methods often use labeled segmentations or diverse signals like text, video, motion, or depth, humans can typically achieve scene decomposition with visual cues alone. Unsupervised object-centric learning, inspired by human abilities and utilizing abundant unlabeled image data, seeks to represent a scene as a composition of distinct objects using only visual information.

Auto-encoding-based frameworks stand at the forefront of object-centric learning approaches~\cite{iodine,burgess2019monet, Engelcke2020GENESIS, Lin2020SPACE,eslami2016attend,slot,singh2022illiterate,lowe2022complexvalued,lowe2023rotating}. Here, the emergence of object-centric representations is enabled due to architectural inductive biases such as the use of bottleneck modules that force the network to prioritize the encoding of salient object features. Their simple design, combined with the ability to operate unsupervised, makes them a standout choice.
One notable advance in this regard 
involves approaches employing `slot'-structured bottlenecks~\cite{slot, singh2022illiterate, dinosaur}. These auto-encoders, characterized by their slot-based architecture, consist of two core components. First is the encoder, responsible for transforming input data into a set of latent vectors referred to as `slots', each intended to represent an individual object within an image. The second is the decoder, burdened with the challenging task of reconstructing the input based on information derived from the extracted slots, guiding the learning of object-centric representations.

To advance this paradigm,
our work introduces SPOT -- a dual-stage strategy designed to elevate object-centric learning so as to more effectively handle complex real-world images, which is one of the pivotal challenges in this domain~\cite{PromisingorElusive}. 
SPOT aims to refine both components of slot-based auto-encoders, enhancing both the encoder's precision in generating object-specific `slots' and the decoder's ability to utilize these slots during reconstruction. 
To that end, it makes two key technical advancements.

\textbf{Improving slot generation through self-training.}
The encoding of object-specific information into the slots is achieved through an iterative attention mechanism that forces slots to compete over image patches. This competition leads to the generation of slot-based attention masks that indicate the association of each image patch with a specific slot~\cite{slot,dinosaur}.  
Besides the encoder, slot-based attention masks are also generated at the decoder side. For autoregressive transformer decoders~\cite{singh2022illiterate, dinosaur}, these are produced by the cross-attention module between each output patch and the extracted slots.
Empirically, we observe that masks produced during decoding demonstrate superior object decomposition, i.e., better object segmentation, compared to those from encoding~\cite{dinosaur}. 
Building on this insight,
we propose a self-training scheme that distills slot-based attention masks from the decoder to the encoder, thereby enhancing the object segmentation information captured by the slots.

\textbf{Enhanced autoregressive decoders with sequence permutations.}
Object-centric models commonly utilize weak slot-wise MLP decoders~\cite{watters2019spatial}. Drawing inspiration from large language models~\cite{brown2020language}, Singh et al~\cite{singh2022illiterate} propose the adoption of more expressive autoregressive transformer decoders in the context of object-centric learning~\cite{singh2022illiterate, singh2022simple}, surpassing the performance of MLP-based counterparts. 
Moreover, multiple studies have underscored the importance of increasing decoder capacity to effectively apply object-centric learning to complex scenes~\cite{dinosaur,chang2022object,wu2023slotformer,singh2023neural,jiang2023object,wu2023slotdiffusion,jabri2023dorsal}.
However, autoregressive transformer models may face overfitting challenges, particularly when accustomed to teacher-forcing training~\cite{williams1989learning},
relying excessively on past ground-truth tokens.
This tendency leads them to neglect slot vectors, offering weaker and less robust supervisory signals for their learning.
To mitigate this, we propose the introduction of sequence permutations, altering the autoregressive transformer's prediction order.
This modification amplifies the role of slot vectors in the reconstruction process, resulting in improved object-centric representations.

In summary, our contributions are threefold:
\begin{enumerate}
    \item 
    We enhance unsupervised object-centric learning in slot-based autoencoders by introducing self-training. This involves distilling slot-attention masks from the initially trained teacher model's decoder to the slot-attention module of the student model, thereby improving precision in generating object-specific slots.
    \item 
    We amplify the role of slot vectors in the reconstruction process by introducing sequence permutations that alter the prediction order of the autoregressive transformer decoder. This modification leads to a more robust supervisory signal for object-centric learning.
    \item 
    Empirical evidence demonstrates the synergistic effectiveness of the above strategies. The combined approach forms the SPOT framework, significantly outperforming prior slot-based autoencoder methods in unsupervised object segmentation, particularly with complex real-world images.

\end{enumerate}

%% file: tex/2_related.tex
\section{Related work}
\label{sec:related}

\parag{Unsupervised object-centric learning}
aims to decompose multi-object scenes into meaningful object entities. Previous studies utilize auto-encoding frameworks~\cite{iodine,burgess2019monet, Engelcke2020GENESIS, Lin2020SPACE,eslami2016attend,slot,singh2022illiterate,lowe2022complexvalued,lowe2023rotating}, with slot-attention bottlenecks~\cite{slot} emerging as a prominent paradigm.
Despite their notability, early object-centric models struggle with complex (real-world) scenes~\cite{PromisingorElusive, karazija2021clevrtex}. To address this, some methods focus on improving precision and stability of the encoder's slot-attention module~\cite{kim2023shepherding, stange2023exploring, kori2023unsupervised, jia2022improving, chang2022object}, employing bi-level optimization~\cite{chang2022object,jia2022improving}, architectural modifications~\cite{kim2023shepherding, kori2023unsupervised}, or regularization losses~\cite{stange2023exploring}. 
Another line of work focuses on designing a decoder that supports good decomposition~\cite{singh2022illiterate, jiang2023object, wu2023slotdiffusion}, where SLATE~\cite{singh2022illiterate} utilizes an autoregressive transformer decoder, and others propose diffusion-based methods~\cite{jiang2023object, wu2023slotdiffusion}. 
DINOSAUR~\cite{dinosaur} uses self-supervised pre-trained features as reconstruction targets, proving more effective in complex scenes.
Additionally, some works extend slot-based auto-encoders to exploit video data~\cite{weis2021benchmarking, kipf2022conditional, singh2022simple,aydemir2023selfsupervised,zadaianchuk2023videosaur,traub2023learning}, motion~\cite{bao2022discovering,bao2023MoTok}, depth~\cite{elsayed2022savi}, or text~\cite{Xu2022groupvit}.
Our work also aims at improving the effectiveness of object centric learning in handling complex real-world data but does so using static images and is thus complementary to the above approaches.

Furthermore, object-centric learning has been explored through contrastive frameworks~\cite{wen2022self,henaff2022object,henaff2021efficient,baldassarre2022towards,lowe2020learning}, primarily for pre-training representations.

\parag{Autoregressive transformer decoders~\cite{autoregressive_transformer}} excel in natural language processing~\cite{xlnet,brown2020language,llama2} and have recently proven effective in computer vision, exemplified by models like iGPT~\cite{chen2020generative}. Notably, SLATE~\cite{singh2022illiterate} and STEVE~\cite{singh2022simple} have applied them to object-centric learning, highlighting their efficacy in handling intricate visual scenes.
However, despite their success, autoregressive transformers face training stability challenges in object-centric learning, potentially due to their high capacity~\cite{dinosaur}. 
They tend to overly depend on past ground-truth tokens, neglecting input from the encoder, as observed in pretraining image encoders with image caption tasks~\cite{tschannen2023image}.
In our analysis, we validate this issue in the context of object-centric learning and propose a simple patch-order permutation strategy, which is very easy to integrate with existing autoregressive transformer models.

\parag{Self-training approaches} refine models using their own predictions on unlabeled data, enhancing the training set. They are widely used in semi-supervised learning for tasks like image classification~\cite{billion2019semisuper, Wei_2021semisuper}, semantic segmentation~\cite{learning_from_future, sem_segm_sefltrain, Yang_2022_CVPR}, and object detection~\cite{meanteachers, Vandeghen_2023_ICCV, minderer_gritsenko_houlsby_2023}, as well as in unsupervised domain adaptation~\cite{UDA1, UDA2, UDA3} and unsupervised localization~\cite{wang2023cut, lost, simeoni2023found}. Additionally, they enhance performance in supervised classification. 
This work introduces self-training to the domain of unsupervised object-centric learning, demonstrating its notable effectiveness in this specific context.

%% file: tex/3_method.tex
\section{Method}
\label{sec:spot}

\begin{figure*}[t]
\centering
\includegraphics[trim={0cm 0cm 0cm 0cm},width=0.715\linewidth]{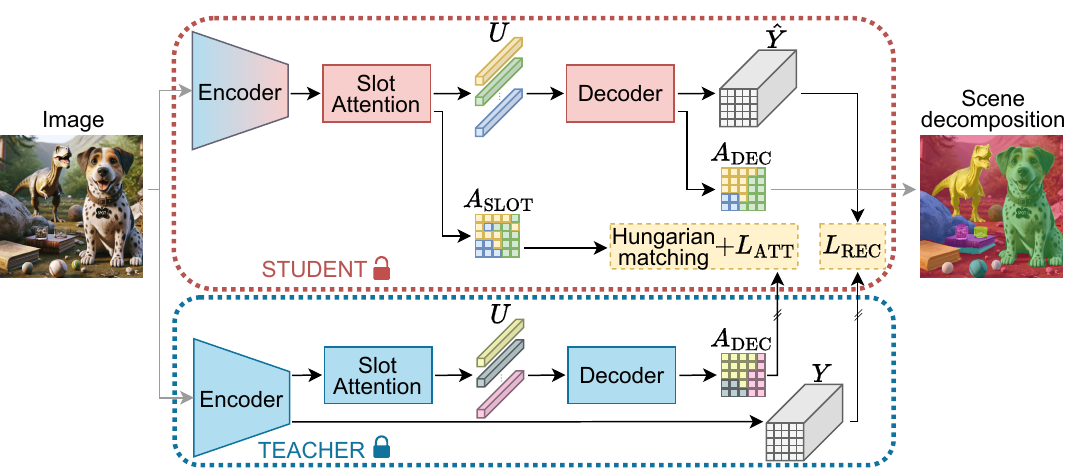}
\vspace{-8pt}
\caption{
\emph{Enhancing unsupervised object-centric learning via self-training}.
Our two-stage approach starts with exclusive training in the initial stage (not depicted) using the reconstruction loss $L_{\mathrm{REC}}$. In the following stage, shown here, a teacher-student framework is applied. The teacher model, trained in the first stage, guides the student model with an additional loss $L_{\mathrm{ATT}}$, distilling attention masks $A_{\mathrm{DEC}}$ from the teacher's decoder to the slot-attention masks $A_{\mathrm{SLOT}}$ in the student's encoder.}
\label{fig:self_training}
\vspace{-14pt}
\end{figure*}

\parag{Slot-based auto-encoders.}
Following the common practice in object-centric learning, we use a slot-based auto-encoding framework~\cite{slot}.
This framework's encoder consists of two modules: 
an image encoder extracting $n$ patch-wise features from image $X$ and a slot-attention module that groups these features into
$k<n$ latent vectors $U=(\vu_1; \dots; \vu_k) \in \real^{k \times d_u}$ referred to as `slots', each representing an object in the image.
The decoder aims to reconstruct a target signal, such as the original image $X$, 
from these slot vectors.
Here, following DINOSAUR~\cite{dinosaur}, we employ self-supervised pre-trained feature encoders (e.g., DINO~\cite{dino}) for instantiating the image encoder as well as extracting the reconstruction targets.
Defining $Y, \hat{Y} \in \real^{n \times d_y}$ as the target features and the predicted reconstructions, respectively, the model is trained by minimizing the reconstruction loss:
\begin{equation}
L_{\mathrm{REC}} = \frac{1}{n \cdot d_y} ||Y - \hat{Y}||_2^2.
\end{equation}
Defining the reconstruction task with high-level features provides a valuable training signal for learning object-centric representations from real-world data~\cite{dinosaur}. 

\parag{Auto-regressive transformer decoders.}
A crucial element in slot-based auto-encoders is the decoder's design~\cite{singh2022simple, singh2022illiterate,jiang2023object,wu2023slotdiffusion}. Here, we employ an autoregressive transformer~\cite{autoregressive_transformer} as our decoder.
Autoregressive transformers predict the feature $\hat{\vy}_i$ at position $i$ based on prior target features $Y_{<i}$ and slots $U$. 
This prediction is jointly performed for all token positions using teacher-forcing\footnote{While joint prediction is typically a training-only practice, in our context, is extended to testing due to the availability of target features.}:
\begin{equation}
    \hat{Y} = \transformerdecoder (Y_{<}; U),
\end{equation}
where $Y_{<}\in \real^{n \times d_y}$ is the decoder's input, consisting of target features $Y$ right-shifted by one position (excluding the last token), with a learnable Beginning-Of-Sentence ($\bos$) token prepended: $Y_{<} = (\vy_{\bos}; \vy_1; \dots; \vy_{n-1})$.  
The decoder is composed of a sequence of transformer blocks~\cite{autoregressive_transformer}, 
each incorporating a causal self-attention layer (to avoid attending to `future' tokens), a patch-to-slot cross-attention layer allowing to utilize information from slot vectors $U$, and a feed-forward layer (see~\autoref{fig:decoder}).

\parag{Objective.}
The primary objective of slot-based auto-encoders is to decompose the input image into its individual objects. 
This is typically evaluated by examining masks linked to each slot, indicating the association of each image patch with a specific slot / object. 
These masks, called slot-attention masks, can be derived from either the slot-attention module or the employed decoder.

In this work, we present SPOT, a novel two-stage training method that enhances object-centric learning on real-world data through self-training (\autoref{sec:self_training}) and sequence permutation in the autoregressive decoder (\autoref{sec:patch_permutations}).

\subsection{Self-training via slot-attention distillation} \label{sec:self_training}

We use the matrix $A \in [0, 1]^{n \times k}$ to denote slot attention masks. Each row of $A$ is a probability distribution in a $k$-dimensional simplex, indicating the assignment of each image patch to $k$ slots.
As previously outlined, we can derive slot-attention masks from two places:

\begin{description}[leftmargin=15pt,topsep=5pt,itemsep=2pt,parsep=0pt,partopsep=0pt]
    \item[Slot-attention module.]    
This module employs an iterative attention-based approach that begins with the initial query slot vectors $\tilde{U}$ to produce the output slot vectors $U$. 
At its core, the module incorporates a modified slot-to-patch cross-attention layer. 
The slot attention module’s matrix $A$
is derived from this layer, specifically from the last iteration's cross-attention map
\begin{equation} \label{eq:slot_masks}
A_{\mathrm{SLOT}} = \softmax\left(\frac{Q_{E}K_{E}\tran}{\sqrt{d_p}}\right)\tran \in \mathbb{R}^{n \times k},
\end{equation}
where $K_{E} \in \mathbb{R}^{n \times d_p}$ are keys computed from the patch-wise features extracted by the image encoder from the input image $X$, and $Q_{E} \in \mathbb{R}^{k \times d_p}$ are queries computed from slot vectors of the previous iteration. 
The $\softmax$ is applied along the slots dimension to enforce competition.
The initial slots vectors $\tilde{U}$ are either independently sampled from a Gaussian distribution~\cite{slot} or are trainable parameters~\cite{jia2022improving}.
    \item[Decoder module.]  
As mentioned earlier, transformer-based decoders incorporate patch-to-slot cross-attention layers for leveraging information from slot vectors $U$.
Here the attention masks $A$, denoted as $A_{\mathrm{DEC}} \in \real^{n \times k}$, are computed as the average (across $H$ heads) of patch-to-slot cross-attention maps from the final transformer layer:
\begin{equation} \label{eq:dec_masks}
A_{\mathrm{DEC}} = \frac{1}{H} \sum_{j=1}^{H} \softmax\left(\frac{Q_j K_j\tran}{\sqrt{d_h}}\right),
\end{equation}
where $K_j \in \real^{k \times d_h}$ are keys computed from the slots $U$, and $Q_j \in \real^{n \times d_h}$ are queries computed from the transformed (from previous layers) decoder input $Y_{<}$. The $\softmax$ is applied along the dimension of slots.
\end{description}

In our empirical analysis of the attention masks from these two modules (see~\autoref{tab:compSPOT} entry (a) in~\autoref{sec:experiments}), we note that the decoder masks exhibit superior performance in grouping patches into object-centric slots.
This observation aligns with the findings reported by Seitzer et al.~\cite{dinosaur}.

\parag{Two-stage training with slot-attention distillation.}
Motivated by this observation, we propose to improve the slot-attention module with a self-training scheme depicted in~\autoref{fig:self_training}.
Our training approach involves two stages. In the initial stage, the model is trained exclusively using the $L_{\mathrm{REC}}$ loss. In the second stage, we employ a teacher-student framework. The pre-trained model serves as the teacher, guiding the training of a new model (referred to as the student) with an additional loss $L_{\mathrm{ATT}}$ that distills attention masks $A_{\mathrm{DEC}}$ from the teacher's decoder, denoted as $A_{T}$, to the attentions masks $A_{\mathrm{SLOT}}$ of the student's encoder, denoted as $A_{S}$.

This distillation enhances the grouping capability of the student's slot-attention module, resulting in improved slot representations. 

\parag{Slot-attention distillation loss $L_{\mathrm{ATT}}$.}
To distill the teacher's attention masks $A_{T}$ to the student, we first convert them from soft to hard-assignment masks by applying row-wise to $A_{T}$ the \texttt{argmax} and \texttt{one-hot} operators: 
\begin{equation}
A'_{T} = \texttt{one-hot}(\texttt{argmax}(A_{T})) \in \{0, 1\}^{n \times k}.   
\end{equation}
Then, we use Hungarian matching~\cite{hungarian} to map the $k$ slots in the teacher's masks $A'_{T}$ with the $k$ slots in the student's masks $A_{S}$, using the IoU between the masks as cost function. The matching process results in the $A''_{T}$ masks.
Finally, we apply the cross-entropy loss between $A''_{T}$ and $A_{S}$:
\begin{equation}
L_{\mathrm{ATT}} = \frac{1}{n} \langle A''_{T}, \log(A_{S}) \rangle_F,
\end{equation}
where $\langle \cdot,\cdot \rangle_F$ denotes the Frobenius inner product. 

The total training loss in the second training stage is
\begin{equation}
L = L_{\mathrm{REC}} + \lambda L_{\mathrm{ATT}},
\end{equation}
where $\lambda$ is the loss weight of the distillation objective.

\parag{Stabilizing image encoder training with $L_{\mathrm{ATT}}$.}
We employ self-supervised pre-trained Vision Transformer~\cite{dosovitskiy2020image} (ViT) models as image encoders, empirically shown to achieve superior object-centric learning results~\cite{dinosaur}.
During the first training stage, it is crucial, as emphasized in~\cite{dinosaur}, to keep the ViT image encoder frozen for training stability to achieve good results.
Our analysis in \autoref{sec:exp_analysis} reveals an additional advantage of the self-training loss: serving as a stabilization factor, it facilitates fine-tuning of ViT in the second training stage, maximizing its learning capacity.
We attribute this stabilizing effect to the self-training loss explicitly guiding the encoder to generate object-specific slot masks, in contrast to the reconstruction loss $L_{\mathrm{REC}}$.

\subsection{Autoregressive transformer decoder with sequence permutations} \label{sec:patch_permutations}

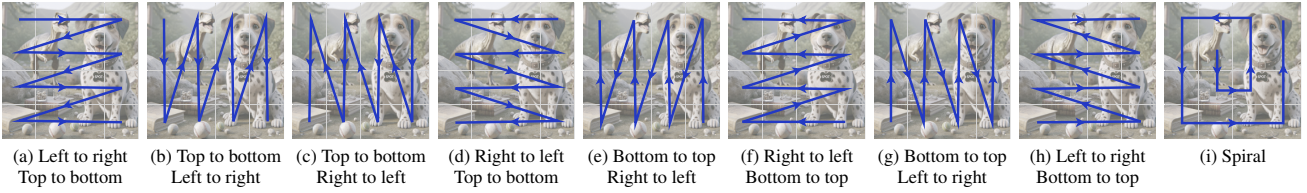
\begin{figure*}[t]
\input{tex/fig_permutations}
\vspace{-8pt}
\caption{\emph{Sequence permutations in SPOT}. The sequence of patches used for autoregressive-based decoder predictions.}
\label{fig:permutations}
\vspace{-13pt}
\end{figure*}
\begin{figure}[t]
\vspace{-14pt}
\input{tex/fig_grad_spot}
\vspace{-8pt}
\caption{
$L_1$ gradients norms for each patch's reconstruction loss with respect to the decoder's input slots (aggregated across all the slots, four decoder blocks, and the entire COCO validation set). 
Subplots show gradients with: (a) default permutation and (b) randomly sampled sequence permutations.
}
\label{fig:mse_spot}
\vspace{-10pt}
\end{figure}
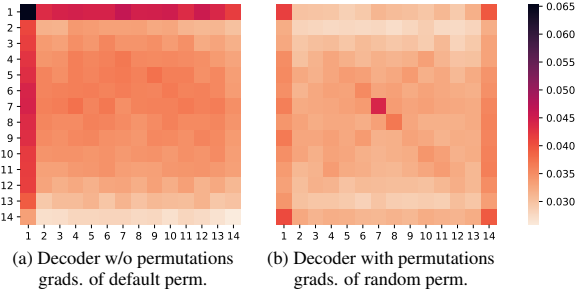

Autoregressive transformer decoders outperform simpler MLPs with spatial-broadcasting decoders~\cite{dinosaur}, thanks to their effectiveness in 
ensuring global consistency in predictions.
However, their high representational capacity, as discussed in~\cite{tschannen2023image}, may limit the encoder's effective learning by diminishing the strength of the supervisory signal backpropagated to it. This is manifested as follows: initial patch tokens heavily rely on slot vector information due to limited context from `past' tokens. 
As decoding progresses, dependence on features of earlier tokens grows, reducing the significance of slot vector information. 
Consequently, later tokens provide a weaker supervisory signal for learning slot vectors in the encoder.
\autoref{fig:mse_spot}(a) illustrates this variability in signal strength based on token location. In this figure, the magnitude of the $L_{\mathrm{REC}}$ loss gradient with respect to slots (averaged over data samples) is presented, highlighting that tokens in the first row and column contribute significantly higher magnitude gradients during backpropagation to slots compared to their later counterparts.

\parag{Autoregressive decoding with permuted sequences.}
To tackle this issue, we propose introducing sequence permutations to alter the order in which the autoregressive transformer predicts, based on a predefined set of permutations. In essence, employing a sequence permutation requires the transformer to make predictions not only in the traditional left-to-right and top-to-bottom patch-order (as depicted in \autoref{fig:permutations}(a)), but also with additional patch-orderings shown in \autoref{fig:permutations}(b-i).
The set of sequence permutations comprises eight ordering variations, each associated with a distinct starting patch and direction (horizontal/vertical), complemented by an additional spiral permutation originating from a patch at the center of the image.

During each training step, we randomly select one sequence permutation from the set to guide predictions by the autoregressive transformer. This introduces variability in token positions, as tokens later in the sequence in the default order may now occupy initial positions in the permuted sequence. Consequently, these tokens must rely more on slot information, offering stronger supervisory signal to the encoder after the permutation.
This improvement is evident in~\autoref{fig:mse_spot}(b), where the magnitudes of the gradients w.r.t. the slots of the model that use sequence permutations in the decoder are less sensitive w.r.t. the token location they are coming from (as opposed to the model without permutations~\autoref{fig:mse_spot}(a)). Thus, the model trained with sequence permutations benefits 
from a more uniform reliance on the slots across all token positions.
Further discussion on the autoregressive transformer is available in the \red{Appendix}~\autoref{sec:discuss_ar_design}.

\begin{figure}[t]
\centering
\includegraphics[trim={0cm 0cm 0cm 0cm},width=0.8\linewidth]{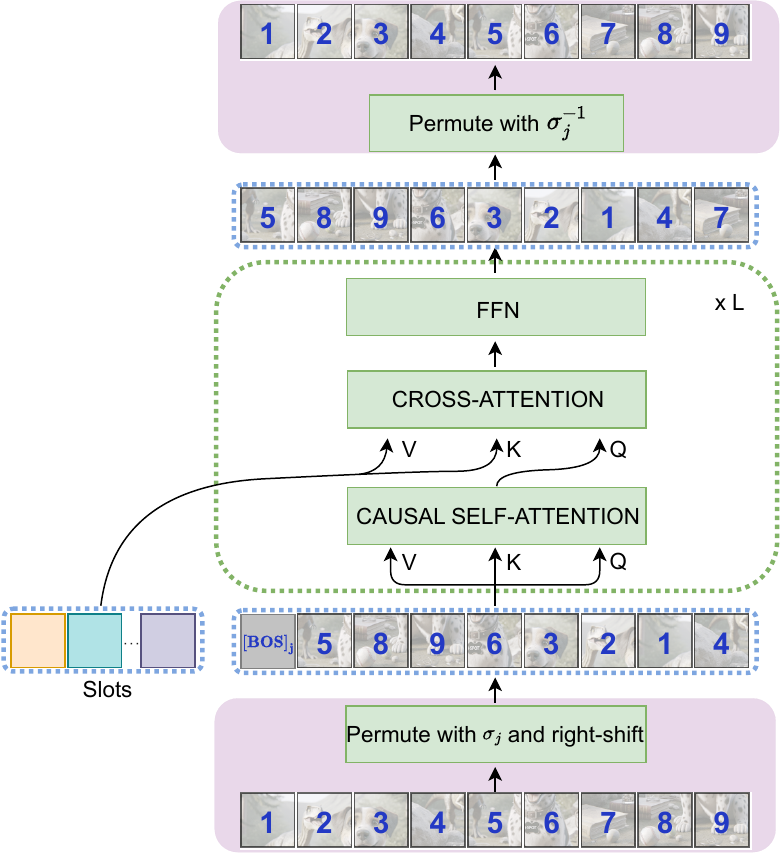}
\vspace{-8pt}
\caption{\emph{Autoregressive (AR) decoding via sequence permutations}. 
Violet boxes indicate differences from typical AR decoder.}
\label{fig:decoder}
\vspace{-10pt}
\end{figure}

We illustrate the workflow of our autoregressive decoder with sequence permutations in~\autoref{fig:decoder}.
To enable the decoder to recognize the current sequence ordering and thus what is the next-patch that it must predict, a different $\bos$ token is used for each different permutation. 
More formally, we define 9 different sequence permutations $\sigma_{j} : \{ 1,\dots, n \} \rightarrow \{ 1,\dots, n \}$, $j=1,\dots,9$.
For each permutation $j$, the permuted input to the autoregressive transformer is 
\begin{equation}
    Y_{< \sigma_{j}} = (\vy_{\bos_{j}}; \vy_{\sigma_{j}(1)}; \dots; \vy_{\sigma_{j}(n-1)}), 
\end{equation}
where $\vy_{\bos_{j}}$ is the $\bos$ token for the $j$ permutation.
The decoder produces predictions $\hat{Y}_{\sigma_{j}}$ for this $j$ permutation 
\begin{equation}
    \hat{Y}_{\sigma_{j}} =  (\vy_{\sigma_{j}(1)}; \dots; \vy_{\sigma_{j}(n)}) = \transformerdecoder (Y_{< \sigma_{j}}; U),
\end{equation}
which are then re-ordered to $\hat{Y}$ using the inverse permutation $\sigma_{j}^{-1}$, followed by applying the $L_{\mathrm{REC}}$ loss.

\input{tex/table_coco_ablations}

Therefore, in our decoder, the sole modifications involve permuting the inputs and outputs and introducing multiple $\bos$ tokens (one per permutation) instead of a single one. Other than that, the core architecture of the decoder remains unchanged.
During training, a single randomly sampled permutation per training step is employed, maintaining the training cost. At inference, we simply use the default permutation (\autoref{fig:permutations}(a)) preserving the same inference cost. Alternatively, computing attention masks $A_{\mathrm{DEC}}$ for each permutation and averaging them (i.e., test-time ensembling using the permutations), as explored in~\autoref{sec:experiments}, yields a slight performance improvement.

\subsection{Incorporating permutations into self-training} 
\label{sec:employ_permutations}

We employ sequence permutation to enhance the transformer decoder in both training stages. 
This improves the 1st-stage decoder, which thus serves as a better teacher for the 2nd self-training stage. 
In the 2nd stage, beyond the student, we also incorporate random sequence permutations in the teacher's decoder when generating the target slot-attention masks for the self-training loss.

For the 1st-stage training, we use as initial slots independently sampled vectors from a Gaussian distribution, following~\cite{slot,dinosaur}. For the 2nd stage, we initialize slots as learnable vectors employing bi-level optimization~\cite{chang2022object,jia2022improving}.

%% file: tex/fig_permutations.tex
\scriptsize
\centering
\setlength{\tabcolsep}{1.4pt}
\begin{tabular}{cccccccccc}
	\fig[.105]{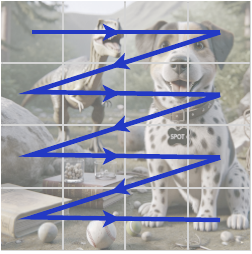}             &
	\fig[.105]{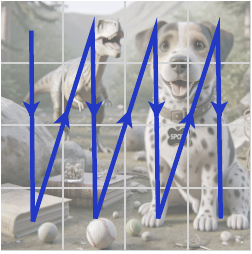}                      &
	\fig[.105]{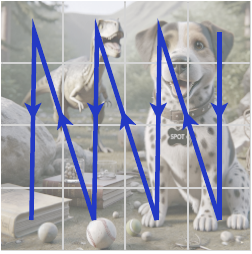}                     &
	\fig[.105]{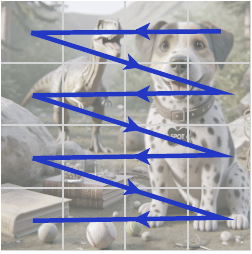}                     &
	\fig[.105]{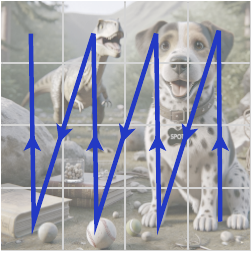}                  &
	\fig[.105]{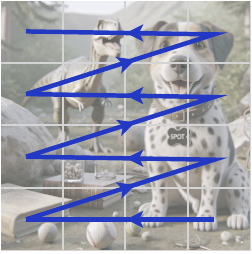}                  &
	\fig[.105]{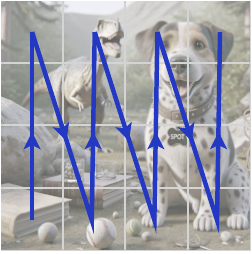}                   &
	\fig[.105]{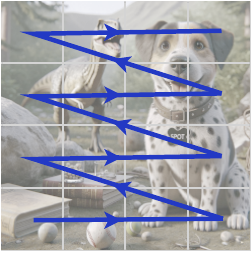}                   &
	\fig[.105]{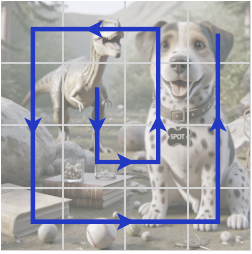}                        \\

	(a) Left to right                                          &
	(b) Top to bottom                                          &
	(c) Top to bottom                                          &
	(d) Right to left                                          &
	(e) Bottom to top                                          &
	(f) Right to left                                          &
	(g) Bottom to top                                          &
 	(h) Left to right                                          &
	(i) Spiral                                                 \\

	Top to bottom                                              &
	Left to right                                              &
	Right to left                                              &
	Top to bottom                                              &
	Right to left                                              &
	Bottom to top                                              &
	Left to right                                              &
 	Bottom to top                                              &
	                                          \\
\end{tabular}

%% file: tex/fig_grad_spot.tex
\scriptsize
\centering
\setlength{\tabcolsep}{1pt}
\begin{tabular}{@{\hspace{-8.0ex}}c@{\hspace{-22.0ex}}c@{\hspace{-50.0ex}}c}
\vspace{-2.5mm}
\fig[0.7]{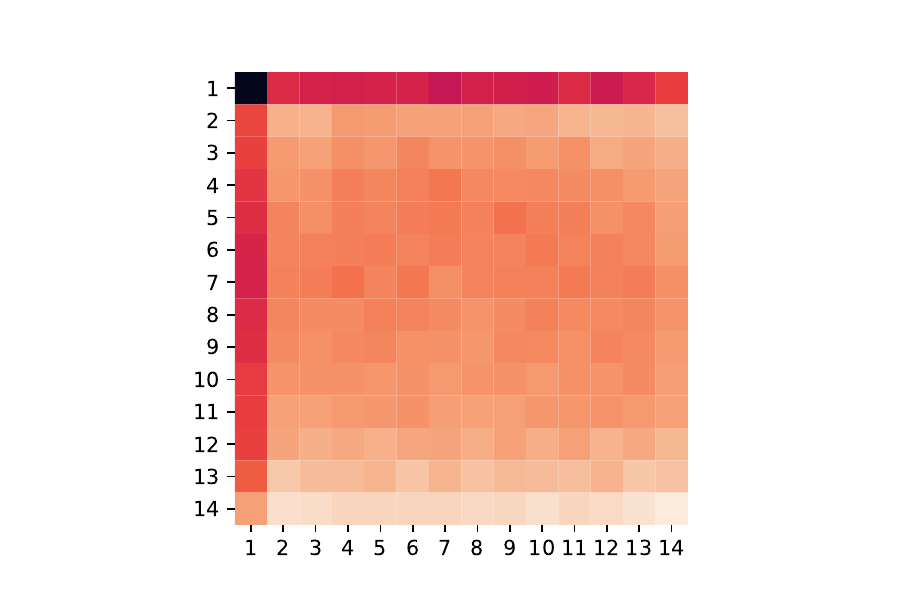} &
\fig[0.7]{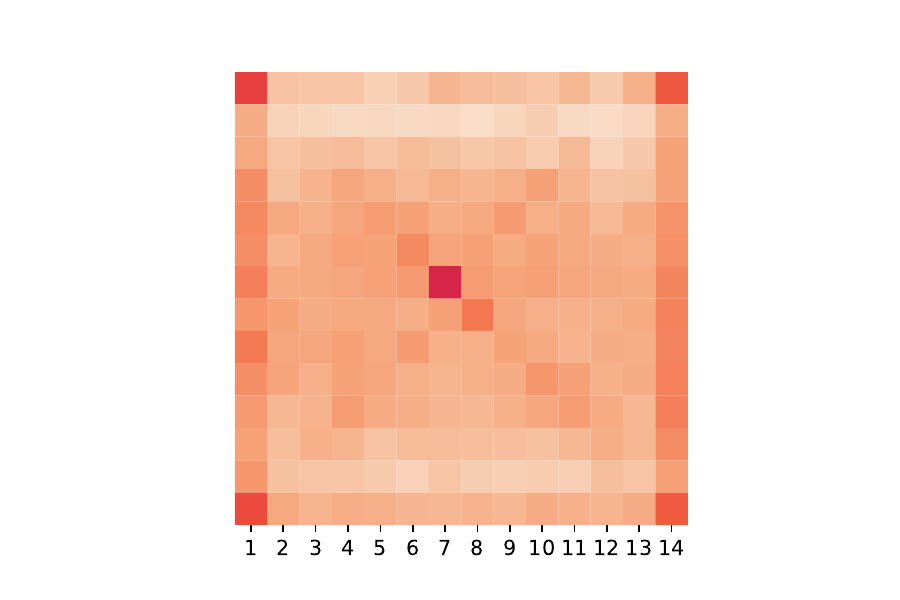} &
\fig[0.7]{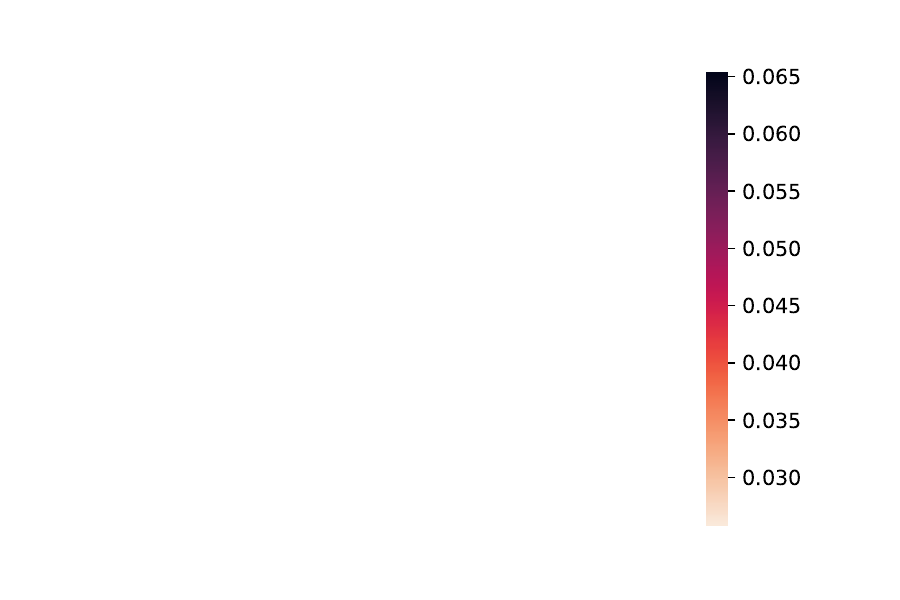}
\\    

(a) Decoder w/o permutations& 

(b) Decoder with permutations&

\\  

grads. of default perm. & 

grads. of random perm. &

\\ 
\end{tabular}

%% file: tex/table_coco_ablations.tex
\begin{table*}[t]
\small
\centering
\setlength{\tabcolsep}{2pt}
\begin{tabular}{lcccccccc|cccc|cccc} \toprule

& \mr{2}{\Th{SP}} & \mr{2}{\Th{ST}} & \mr{2}{\Th{ENS}} & \mr{2}{\Th{Epochs}} & \mc{4}{\decoder}  & \mc{4}{\encoder}& \mc{4}{\maxdecenc} \\\cmidrule{6-17}

 & & & & & \mboi & \mboc & \fgiou & \fgari & \mboi & \mboc & \fgiou & \fgari & \mboi & \mboc & \fgiou & \fgari \\\midrule

(a) & & & & 50 & 32.0{\tiny $\pm$0.1} & 41.4{\tiny $\pm$0.3} & 30.0{\tiny $\pm$0.1} & 32.3{\tiny $\pm$0.6} & 30.0{\tiny $\pm$0.2} & 38.9{\tiny $\pm$0.4}& 28.1{\tiny $\pm$0.2} & 29.5{\tiny $\pm$0.6} &32.0{\tiny $\pm$0.1} & 41.4{\tiny $\pm$0.3} & 30.0{\tiny $\pm$0.1} & 32.3{\tiny $\pm$0.6} \\

(b) & \cmark & & & 50 & {32.7\tiny $\pm$0.2} & 40.9{\tiny $\pm$0.5} & 30.8{\tiny $\pm$0.1} & 35.6{\tiny $\pm$0.5}  & 31.1{\tiny $\pm$0.1} & 38.9{\tiny $\pm$0.3} & 29.4{\tiny $\pm$0.1} & 33.4{\tiny $\pm$0.4} &{32.7\tiny $\pm$0.2} & 40.9{\tiny $\pm$0.5} & 30.8{\tiny $\pm$0.1} & 35.6{\tiny $\pm$0.5} \\

(c) & \cmark & & \cmark & 50 & 32.9{\tiny $\pm$0.2} & 41.2{\tiny $\pm$0.5} & 31.0{\tiny $\pm$0.1} & 36.0{\tiny $\pm$0.5} & 31.1{\tiny $\pm$0.1} & 38.9{\tiny $\pm$0.3} & 29.4{\tiny $\pm$0.1} & 33.4{\tiny $\pm$0.4} & 32.9{\tiny $\pm$0.2} & 41.2{\tiny $\pm$0.5} & 31.0{\tiny $\pm$0.1} & 36.0{\tiny $\pm$0.5} \\ \midrule

(d) & & & & 100 & {32.3\tiny $\pm$0.3} & {42.1\tiny $\pm$0.2} & 30.2{\tiny $\pm$0.3} & {31.8\tiny $\pm$0.9} & {30.5\tiny $\pm$0.2} & {39.8\tiny $\pm$0.1} & 28.5{\tiny $\pm$0.3} & {30.0\tiny $\pm$0.7} &{32.3\tiny $\pm$0.3} & {42.1\tiny $\pm$0.2} & 30.2{\tiny $\pm$0.3} & {31.8\tiny $\pm$0.9} \\ 

(e) & & \cmark & & 50+50 & 30.1{\tiny $\pm$0.5} & 38.2{\tiny $\pm$1.0} & 28.3{\tiny $\pm$0.4} & 22.5{\tiny $\pm$1.6} & 33.2{\tiny $\pm$0.1} & 43.6{\tiny $\pm$0.2} & 31.1{\tiny $\pm$0.0} & 34.4{\tiny $\pm$0.5} &33.2{\tiny $\pm$0.1} & 43.6{\tiny $\pm$0.2} & 31.1{\tiny $\pm$0.0} & 34.4{\tiny $\pm$0.5} \\

\rowcolor{TableColor} (f) & \cmark & \cmark & & 50+50 & \bf{34.7{\tiny $\pm$0.1}} & \bf{44.3{\tiny $\pm$0.3}} & \bf{32.7{\tiny $\pm$0.1}} & \bf{36.6{\tiny $\pm$0.3}} & \bf{33.7{\tiny $\pm$0.1}} & \bf{43.1{\tiny $\pm$0.4}} & \bf{31.8{\tiny $\pm$0.1}} & \bf{37.8{\tiny $\pm$0.5}} &\bf{34.7{\tiny $\pm$0.1}} & \bf{44.3{\tiny $\pm$0.3}} & \bf{32.7{\tiny $\pm$0.1}} & \bf{37.8{\tiny $\pm$0.5}} \\

\rowcolor{TableColor} (g) & \cmark & \cmark & \cmark & 50+50 & \bf{35.0{\tiny $\pm$0.1}} & \bf{44.7{\tiny $\pm$0.3}} & \bf{33.0{\tiny $\pm$0.1}} & \bf{37.0{\tiny $\pm$0.2}} & \bf{33.7{\tiny $\pm$0.1}} & \bf{43.1{\tiny $\pm$0.4}} & \bf{31.8{\tiny $\pm$0.1}} & \bf{37.8{\tiny $\pm$0.5}} &\bf{35.0{\tiny $\pm$0.1}} & \bf{44.7{\tiny $\pm$0.3}} & \bf{33.0{\tiny $\pm$0.1}} & \bf{37.8{\tiny $\pm$0.5}} \\
\bottomrule
\end{tabular}
\vspace{-8pt}
\caption{\emph{Ablation study on COCO}. Results for slot masks generated by \decoder, \encoder, and their max (\maxdecenc) using mean and std over 3 seeds. \Th{SP}: sequence permutation, \Th{ST}: self-training, \Th{ENS}: ensembling of nine permutations at test-time.}
\label{tab:compSPOT}
\vspace{-10pt}
\end{table*}

%% file: tex/4_experiments.tex
\section{Experiments}
\label{sec:experiments}

\input{tex/table_self_training_analysis}

\subsection{Setup}

\parag{Datasets.}
We utilized the MS COCO 2017~\cite{coco} dataset for its diverse collection of real-world images, each featuring multiple co-occurring objects. This dataset poses a significant challenge for object-centric learning models due to the complexity of the scenes. We also considered the PASCAL VOC 2012~\cite{voc} dataset, which comprises images often containing a single or just a few salient objects, offering a comparatively more straightforward evaluation. Furthermore, we used the synthetic datasets MOVi-C and MOVi-E~\cite{movi}, which contain approximately 1000 realistic 3D-scanned objects. MOVi-C includes scenes with 3-10 objects, whereas MOVi-E contains scenes with 11-23 objects per scene. Although MOVi-C/E are originally video-based, we adapt them by selecting random frames following \cite{dinosaur}.

\parag{Metrics}
To assess object-centric learning, we use Mean Best Overlap at the instance (\mboi) and class (\mboc) levels, Foreground Adjusted Rand index (\fgari), and mean Intersection over Union (\miou).
\mboi identifies the best overlap for each ground truth mask, while \fgiou employs Hungarian matching for a one-to-one correspondence between predicted and ground truth segments.
Our main focus is on \mbo and \fgiou metrics because they consider background pixels, comprehensively evaluating how closely masks fit around objects. In contrast, \fgari, a cluster similarity metric, exclusively focuses on foreground pixels, potentially giving a misleading impression of segmentation quality by ignoring the fidelity of predicted masks while also promoting over-segmentation. While we report \fgari, it is not our primary focus, aligning with concerns raised in other studies~\cite{wu2023slotdiffusion,Engelcke2020GENESIS, karazija2021clevrtex, monnier2021dtisprites}. We provide a more detailed discussion about \fgari unreliability in the~\autoref{sec:unreliable_fgari}.

\parag{Implementation details.}
We employ the Adam optimizer~\citep{kingma2015adam} with $\beta_1 = 0.9$, $\beta_2 = 0.999$, no weight decay, and a batch size of 64. 
For each training stage on COCO and PASCAL, we use 50 and 560 training epochs, respectively. For MOVi-C/E experiments, we use 65 and 30 epochs for the first and second stages, respectively.

We implement our SPOT models using ViT-B/16~\cite{dosovitskiy2020image} for the encoder (by default initialized with DINO~\cite{dino}) and 4 transformer layers in our decoder.
Unless stated otherwise, loss weight $\lambda$ is 0.005. 
Following~\cite{dinosaur}, on COCO, PASCAL, MOVi-C, and MOVi-E we use 7, 6, 11, and 24 slots, respectively. 
All models are trained on a single GPU with 24 Gbytes. 
We provide learning rate schedules and further implementation details in the~\autoref{sec:more_details}.

\subsection{Analysis} \label{sec:exp_analysis}

In this section, we analyze our approach, emphasizing the impact of self-training and sequence-permutation in the autoregressive decoder, along with related design choices. 
Our primary experimentation focuses on the challenging COCO dataset, complemented by evaluations on MOVi-C.

\parag{(A) Sequence-permutation impact.} \autoref{tab:compSPOT} evaluates the influence of sequence-permutation with and without self-training. Without self-training, sequence permutations (\autoref{tab:compSPOT} (a)$\rightarrow$(b)) enhances instance-specific metrics (\mboi, \fgiou, \fgari) on both the decoder and slot-attention. 
The class-specific metric \mboc remains stable (within std). 

When self-training is enabled, the significance of sequence permutations becomes even more evident (\autoref{tab:compSPOT} (e)$\rightarrow$(f)). 
Omitting them results in a notable performance drop across all metrics, especially impacting decoder-specific ones (\decoder cols.). 
Notably, without sequence-permutations, self-training boosts the slot-attention module (\encoder cols.) but hampers decoder performance (\decoder cols.). This discrepancy highlights the autoregressive decoder's susceptibility to neglecting slot input during reconstruction tasks. An overfitting behavior of this type may be attributed to the accelerated learning dynamics of slot vectors during training, caused by the self-training loss $L_{\mathrm{ATT}}$ applied to them, making it challenging for the autoregressive decoder to effectively leverage them. 
Sequence permutations play a crucial role in compelling the decoder to prioritize slots, underscoring their importance in autoregressive decoders for slot-centric learning.

\textit{\textbf{Test-time ensembling of permutations.}}
Comparing models (b) to (c) and (f) to (g) in \autoref{tab:compSPOT}, we observe a slight performance gain by employing test-time ensembling of permutations.
However, this comes at the expense of increased inference time. Nonetheless, our method demonstrates very robust performance even without test-time ensembling, emphasizing that the core influence of sequence-permutation lies in enhancing the effectiveness and stability of training slot-based auto-encoders with autoregressive decoders.

\textit{\textbf{Comparison to other AR training approaches.}}
Motivated by CapPa~\cite{tschannen2023image}, we explored employing a training procedure that switches between autoregressive and parallel non-autoregressive decoding\footnote{Here, given as input position embeddings, the transformer predicts at all positions in parallel without causal masking in self-attention.}, with the latter being used at 25\% of training steps, as an alternative to sequence permutations for enhancing the autoregressive decoder. Our findings, outlined in~\autoref{tab:more_decoders}, reveal that, in the context of object-centric learning, parallel decoding during training does not help, 
unlike our sequence permutation approach.

\input{tex/table_more_decoders}

Last, the positive impact of sequence permutations is also demonstrated in MOVi-C (results in the appendix~\autoref{sec:movic_ablation}).

\parag{(B) Self-training impact.} 
Comparing entries (f) to (b) in \autoref{tab:compSPOT}, 
we see that self-training yields significant gains. 
It also helps the max performance (columns \maxdecenc) in the case of not using sequence permutation. 

\input{tex/table_mlp_decoder}

\input{tex/table_comparison_prior_work_horizontal_without_ari}

\textit{\textbf{Autoregressive and MLP-based decoders.}}
As discussed earlier, when it comes to autoregressive decoder performance, optimal results are obtained when self-training is paired with sequence-permutation, which is crucial for preventing the decoder from ignoring slot information and overfitting.
Notably, our self-distillation scheme also extends beyond autoregressive decoders. To showcase this, we apply it with MLP-based decoders where it brings significant performance improvements as demonstrated in \autoref{tab:decoder}.

In~\autoref{tab:self_training_analysis}, section (a) illustrates the sensitivity of the self-distillation scheme to the loss weight $\lambda$, with $\lambda$=0.005 producing optimal results. Performance remains relatively stable for loss weights around this value.

In section (b) of~\autoref{tab:self_training_analysis}, we investigate the impact of extracting the teacher's slot-mask $A_{T}$ with a standard or random permutation during the self-distillation scheme. This design choice does not appear to have a significant impact, with a random permutation being our default choice.

Finally, in section (c) of~\autoref{tab:self_training_analysis}, 
we highlight an additional benefit derived from the self-training stage, particularly in the successful fine-tuning of the image encoder during that stage.
We stress that, without the self-training loss, allowing the image encoder to be fine-tuned becomes unstable, yielding subpar results (\autoref{tab:self_training_analysis} (a) for $\lambda=0$ in $L_{\mathrm{ATT}}$). 
This observation aligns with Seitzer et al.'s findings~\cite{dinosaur}. 
The self-training loss acts as a stabilizing factor, enabling further utilization of the pre-trained image encoder. We provide further analysis of the training stability of the image encoder in appendix~\autoref{sec:finetuning_further_analysis}.

\subsection{Comparison with object-centric methods}

\begin{figure}[tb]
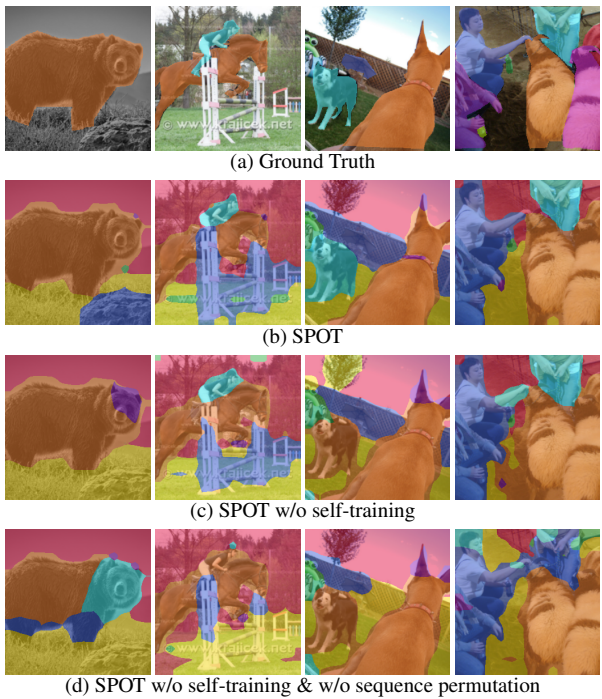

    \centering
    \include{tex/fig_coco_examples}
    \vspace{-15pt}
    \caption{Example results on COCO 2017, using 7 slots.
    }
    \label{fig:results-object-discovery-real}
    \vspace{-10pt}
\end{figure}

\parag{Method comparison.}
In~\autoref{tab:benchmark_v2}, we compare our object-centric learning method SPOT with prior approaches across the MOVi-C, MOVi-E, PASCAL, and COCO datasets using \mboi and, when applicable, \mboc metrics. More detailed results, including the \fgari metric are provided in the~\autoref{sec:more_benchmarking}. 
Our method outperforms others in all scenarios except PASCAL's \mboi, where it ranks second to SlotDiffusion~\cite{wu2023slotdiffusion}.
Notably, on the demanding COCO dataset, our approach excels, surpassing the prior state-of-the-art by 2.7 and 5.9 points in \mboi and \mboc metrics, respectively. This underscores its superiority in unsupervised object-centric learning with real-world data.

In the \red{Appendix}~\autoref{sec:boqsa_other}, we provide additional results about the impact of bi-level optimized queries~\cite{jia2022improving} on DINOSAUR~\cite{dinosaur} and SlotDiffusion~\cite{wu2023slotdiffusion} frameworks. Additionally, in the \red{Appendix}~\autoref{sec:other_encoders}, we examine SPOT with other pre-trained image features using MoCo-v3~\cite{chen2021empirical} and MAE~\cite{he2022masked} encoders.

\subsection{Qualitative results}

In ~\autoref{fig:results-object-discovery-real}, we present the effects of self-training and sequence permutation. This combination effectively mitigates over-segmentation issues while preserving a high degree of detailed segmentation.
More examples of the efficacy of SPOT are provided in ~\autoref{fig:teaser}, showcasing its robust performance across various images.

%% file: tex/table_self_training_analysis.tex
\begin{table*}[!t]\centering
\small
\centering
\resizebox{0.75\linewidth}{!}{
\centering
\begin{tabular}{ccccc|cc|cc} \toprule
\mc{5}{(a) \Th{Loss Weight $\lambda$}} & \mc{2}{(b) \Th{Permutations}} & \mc{2}{(c) \Th{Enc. fine-tuning}} \\ \midrule
0 & 0.002    & \cellcolor{TableColor}0.005        & 0.01   & 0.02    & \Th{Default} & \cellcolor{TableColor}\Th{Random} & \xmark & \cellcolor{TableColor}\cmark\\ \midrule

 30.7{\tiny $\pm$2.2} & 34.6{\tiny $\pm$0.3} & \cellcolor{TableColor}\bf{34.7{\tiny $\pm$0.1}} & 34.1{\tiny $\pm$0.5} & 33.4{\tiny $\pm$1.5} & 34.6{\tiny $\pm$0.2} & \cellcolor{TableColor}\bf{34.7{\tiny $\pm$0.1}} & 32.4{\tiny $\pm$0.2} & \cellcolor{TableColor}\bf{34.7{\tiny $\pm$0.1}} \\ \bottomrule
\end{tabular}}
\vspace{-8pt}
\caption{\emph{Analysis of self-training hyper-parameters on COCO}. 
Results are the mean and standard deviation of the decoder's \mboi over 3 seeds.
Section (a) studies the impact of the loss weight $\lambda$, section (b) the impact of using random or the default permutation for generating the target mask $A_{T}$ with the teacher model, and section (c) the impact of fine-tuning the image encoder during the self-training stage.
}
\label{tab:self_training_analysis}
\vspace{-15.8pt}
\end{table*}

%% file: tex/table_more_decoders.tex
\begin{table}[t]
\small
\centering
\setlength{\tabcolsep}{4pt}
\begin{tabular}{l|cccc} \toprule
 \Th{Decoder}& \mboi & \miou & \fgari \\ \midrule
 \Th{Transformer}   & 32.0 & 30.0  & 32.3 \\
 \Th{Transformer w/ PA} & 27.8 & 26.5  & 35.3 \\ 
\rowcolor{TableColor} \Th{Transformer w/ SP} & \bf{32.7} & \bf{30.8} & \bf{35.6} \\ 
\bottomrule
\end{tabular}
\vspace{-8pt}
\caption{\emph{Autoregressive decoder comparison on COCO}.
Evaluation metrics for slot masks generated by the autoregressive transformer decoder trained conventionally (Transformer), with sequence permutation during training (\Th{Transformer w/ SP}), or parallel prediction for 25\% of training iterations (\Th{Transformer w/ PA}), as discussed in~\cite{tschannen2023image}. No self-training is employed.}
\label{tab:more_decoders}
\vspace{-10pt}
\end{table}

%% file: tex/table_mlp_decoder.tex
\begin{table}[t]
\small
\centering
\setlength{\tabcolsep}{4pt}
\begin{tabular}{l|c|ccccc} \toprule
 \Th{Decoder}& \Th{ST} & \mboi & \mboc & \miou & \fgari \\ \midrule
\mr{2}{MLP} & \xmark & 26.7 & 30.3 & 25.6 & 38.7 \\ 
 & \cellcolor{TableColor}\cmark & \cellcolor{TableColor}\bf{28.4} & \cellcolor{TableColor}\bf{32.4} & \cellcolor{TableColor}\bf{27.0} & \cellcolor{TableColor}\bf{42.5} \\
\bottomrule
\end{tabular}
\vspace{-8pt}
\caption{\emph{Self-training with MLP decoder on COCO}. 
Evaluation metrics for slot masks generated by the decoder. \Th{ST}: self-training.
}
\label{tab:decoder}
\vspace{-10pt}
\end{table}

%% file: tex/table_comparison_prior_work_horizontal_without_ari.tex
\begin{table*}[t]
\small
\centering
\setlength{\tabcolsep}{6pt}
\begin{tabular}{lcc|cc|cc|cc} \toprule
\mr{2}{\Th{Method}}  & \mc{2}{\Th{COCO}} & \mc{2}{\Th{PASCAL}} & \mc{2}{\Th{MOVi-C}} & \mc{2}{\Th{MOVi-E}} \\ \cmidrule{2-9}
& \mboi & \mboc  & \mboi & \mboc  & \mboi & \miou & \mboi & \miou\\\midrule 

SA~\cite{slot}$^{\dagger}$  & 17.2\hspace{11.5pt} & 19.2\hspace{11.5pt}  & 24.6\hspace{11.5pt} & 24.9\hspace{11.5pt}  & 26.2{\tiny $\pm$1.0} & - & 24.0{\tiny $\pm$1.2} & - \\ 

SLASH~\cite{kim2023shepherding}  & - & -  & - & -  & - & 27.7{\tiny $\pm$5.9} & - & - \\ 

SLATE~\cite{singh2022illiterate}$^{\dagger}$  & 29.1\hspace{11.5pt} & 33.6\hspace{11.5pt}  & 35.9\hspace{11.5pt} & 41.5\hspace{11.5pt}  & 
39.4{\tiny $\pm$0.8}
& 37.8{\tiny $\pm$0.7}& 
30.2{\tiny $\pm$1.7}
& 28.6{\tiny $\pm$1.7} \\ 

CAE~\cite{lowe2022complexvalued}$^{\dagger}$  & - & -  & 32.9{\tiny $\pm$0.9} & 37.4{\tiny $\pm$1.0}  & - & - & - & - \\ 

DINOSAUR~\cite{dinosaur}         & 32.3{\tiny $\pm$0.4} & 38.8{\tiny $\pm$0.4}  & 44.0{\tiny $\pm$1.9} & 51.2{\tiny $\pm$1.9}  & 42.4\hspace{11.5pt} & - & - & - \\ 

DINOSAUR-MLP~\cite{dinosaur}   & 27.7{\tiny $\pm$0.2} & 30.9{\tiny $\pm$0.2}  & 39.5{\tiny $\pm$0.1} & 40.9{\tiny $\pm$0.1}  & 39.1{\tiny $\pm$0.2} & - & 35.5{\tiny $\pm$0.2} & - \\ 

Rotating Features~\cite{lowe2023rotating}  & - & -  & 40.7{\tiny $\pm$0.1} & 46.0{\tiny $\pm$0.1}  & - & - & - & - \\ 

SlotDiffusion~\cite{wu2023slotdiffusion}   & 31.0\hspace{11.5pt} & 35.0\hspace{11.5pt}  & \bf{50.4}\hspace{11.5pt} & \underline{55.3}\hspace{11.5pt}  & - & - & 30.2\hspace{11.5pt} & 30.2\hspace{11.5pt} \\ 

(Stable-)LSD~\cite{jiang2023object}       & 30.4\hspace{11.5pt} & -  & - & - & 45.6{\tiny $\pm$0.8} & 44.2{\tiny $\pm$0.9} & 39.0{\tiny $\pm$0.5} & 37.6{\tiny $\pm$0.5} \\
\rowcolor{TableColor}{SPOT w/o ENS (ours)}  & \underline{34.7{\tiny $\pm$0.1}} & \underline{44.3{\tiny $\pm$0.3}} & 48.1{\tiny $\pm$0.4} & \underline{{55.3{\tiny $\pm$0.4}}} & \underline{47.0{\tiny $\pm$1.2}} & \underline{46.4{\tiny $\pm$1.2}} & \underline{39.9{\tiny $\pm$1.1}} & \underline{39.0{\tiny $\pm$1.1}} \\

\rowcolor{TableColor}{SPOT w/ ENS (ours)} & \bf{35.0{\tiny $\pm$0.1}} & \bf{44.7{\tiny $\pm$0.3}} & \underline{{48.3{\tiny $\pm$0.4}}} & \bf{55.6{\tiny $\pm$0.4}} & \bf{47.3{\tiny $\pm$1.2}} & \bf{46.7{\tiny $\pm$1.3}} & \bf{40.1{\tiny $\pm$1.2}} & \bf{39.3{\tiny $\pm$1.2}} \\
\bottomrule
\end{tabular}
\vspace{-8pt}
\caption{\emph{Comparison with object-centric methods on COCO, PASCAL, MOVi-C and MOVi-E datasets}.
SPOT results are the mean and std over 3 seeds. 
DINOSAUR uses an autoregressive decoder and DINO~\cite{dino} ViT encoder (ViT-B/16 for PASCAL and MOVi-C, ViT-S/8 for COCO).
DINOSAUR-MLP uses an MLP decoder and DINO ViT encoder (ViT-B/16 for COCO and PASCAL, ViT-S/8 for MOVi-C/E).
$^{\dagger}$: COCO and PASCAL results of SA and SLATE are from \cite{wu2023slotdiffusion}, MOVi-C/E results are from \cite{dinosaur} for SA and from \cite{jiang2023object} for SLATE, PASCAL results of CAE are from \cite{lowe2023rotating}.
}
\label{tab:benchmark_v2}
\vspace{-10pt}
\end{table*}

%% file: tex/fig_coco_examples.tex
{
\footnotesize
\centering
\newcommand{\myg}[1]{\includegraphics[width=0.11\textwidth,valign=c]{#1}}
\setlength{\tabcolsep}{1pt}
\begin{tabular}{@{}cccc@{}}
   
    \myg{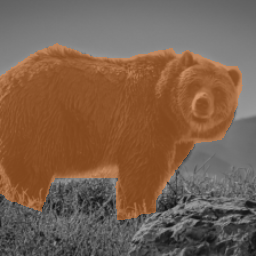} & 
    \myg{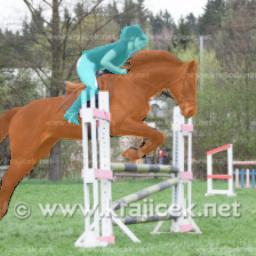} & 
    \myg{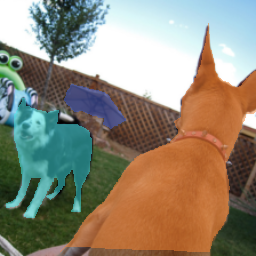} &
    \myg{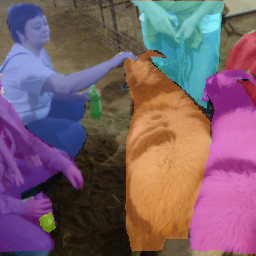} \\
    \mc{4}{(a) Ground Truth}\\

    \mc{4}{\vspace{-2.1ex}}\\
    \myg{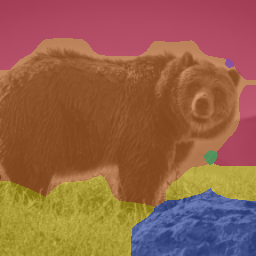} & 
    \myg{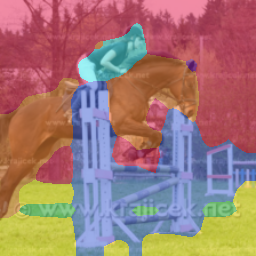} & 
    \myg{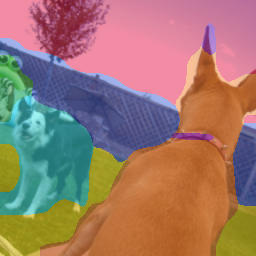} &
    \myg{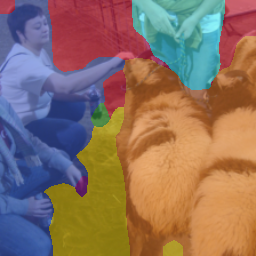} \\

    \mc{4}{(b) SPOT
    }\\

    \mc{4}{\vspace{-2.1ex}}\\

    \myg{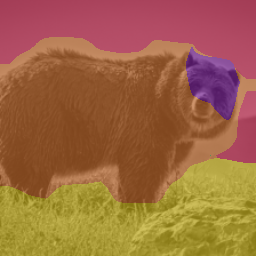} & 
    \myg{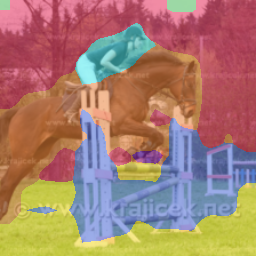} & 
    \myg{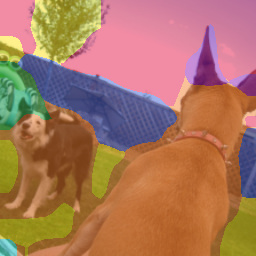} &
    \myg{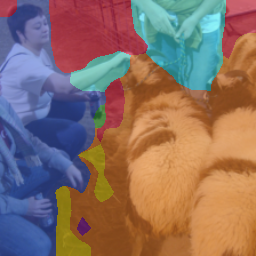} \\

    \mc{4}{(c) SPOT w/o self-training
    }\\ 

    \mc{4}{\vspace{-2.1ex}}\\
    \myg{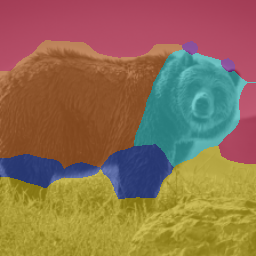} & 
    \myg{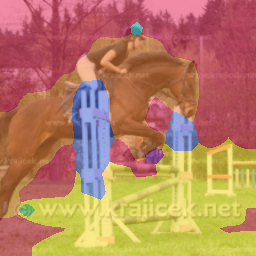} & 
    \myg{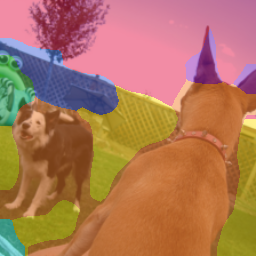} &
    \myg{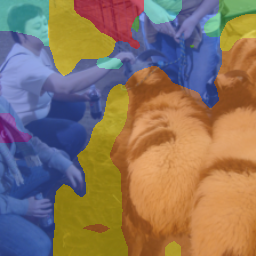} \\

    \mc{4}{(d) SPOT w/o self-training \& w/o sequence permutation}\\ 

\end{tabular}
}

%% file: tex/5_conclusion.tex
\section{Conclusion}
\label{sec:conclusion}

In conclusion, SPOT advances unsupervised object-centric learning for real-world images by enhancing slot-based auto-encoders through two key strategies. Firstly, a self-training scheme uses decoder-generated attention masks to improve slot attention in the encoder. Secondly, a novel patch-order permutation strategy for autoregressive transformers boosts the decoder's performance without additional training cost. 
The synergistic application of these strategies enables SPOT to achieve state-of-the-art results in real-world object-centric learning.

We note that sequence permutation decoding might be more broadly beneficial to other computer vision tasks employing autoregressive decoders.

We discuss the limitations of our method and directions for future work in the \red{Appendix}~\autoref{sec:limitations}.

%% file: tex/acknowledgements.tex
\paragraph{Acknowledgements}

This research work was supported by the Hellenic Foundation for Research and Innovation (HFRI) and the General Secretariat of Research and Innovation (GSRI) under the 4th Call for H.F.R.I. Scholarships to PhD Candidates (grant: 11252). It was also partially supported by the RAMONES and iToBos EU Horizon 2020 projects under grants 101017808 and 965221, respectively. NTUA thanks NVIDIA for the support with the donation of GPU hardware.

%% file: tex/supplementary.tex
\section{Discussion about \fgari unreliability}
\label{sec:unreliable_fgari}

\begin{figure}[tb]
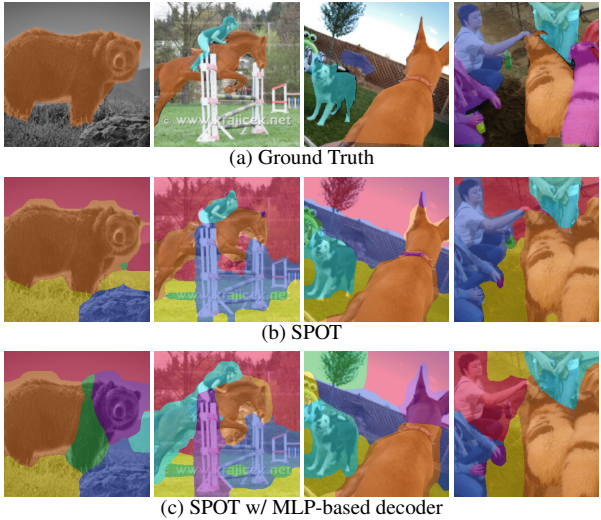

    \centering
    \include{tex/fig_coco_mlp}
    \vspace{-15pt}
    \caption{Example results on COCO 2017, using 7 slots.
    }
    \label{fig:spot_mlp_coco}
    \vspace{-10pt}
\end{figure}

\input{tex/table_comparison_prior_work}


\fgari has been a common metric in assessing predicted object masks against ground-truth segmentation in previous works. However, several recent works have raised concerns about its reliability~\cite{wu2023slotdiffusion,Engelcke2020GENESIS, karazija2021clevrtex, monnier2021dtisprites, dinosaur}. 
Notably, FG-ARI is criticized for its unreliability, which may favor either over-segmentation~\cite{Engelcke2020GENESIS, monnier2021dtisprites} or under-segmentation. Additionally, the fact that it ignores background pixels makes it unable to probe a model's effectiveness in object segmentation~\cite{karazija2021clevrtex,monnier2021dtisprites}.

Consequently, \fgari can be misleading in assessing segmentation quality. We illustrate this in \autoref{tab:further_benchmark}, where we show that we can achieve the highest 49.9 \fgari score in the Pascal dataset \emph{by trivially assigning all pixels to a single slot-mask (1-block mask)}. This highlights the metric's unreliability, particularly in datasets with scenes featuring few or single objects.

To further examine \fgari, \autoref{fig:spot_mlp_coco} compares qualitative results of SPOT with an autoregressive transformer decoder (the standard setup) and SPOT with an MLP-based decoder. 
While the standard SPOT setup outperforms SPOT with an MLP decoder in \mboi, \mboc, and \miou metrics, SPOT with MLP achieves a higher \fgari score (42.5 vs. 37.8) (see Table 1 and Table 4 in the main paper). However, the qualitative results in \autoref{fig:spot_mlp_coco} indicate that SPOT with MLP is distinctly inferior to the standard SPOT. This is particularly evident in the first three images, showing notable over-segmentation (as seen with the bear, horse, and dog) and incorrect object grouping (such as the horse and rider).

These observations emphasize the inadequacy of \fgari in measuring the segmentation quality of predicted object masks. Unsupervised object-centric methods should place greater reliance on \mbo and \miou metrics.

\input{tex/table_movic_ablations}

\input{tex/table_self_training_further_analysis}

\section{Additional experimental results}
\label{sec:more_benchmarking}
\subsection{Comparison with prior object-centric methods}

In Table \ref{tab:further_benchmark}, we present extended benchmark results, including the mean Intersection over Union (\miou) and Foreground Adjusted Rand index (\fgari) across all datasets. We show results for SPOT on \fgari derived from both decoder and slot encoder attention masks, offering a more comprehensive view of its capabilities.

SPOT outperforms other methods in \mboi, \mboc and \miou across all datasets, except for the \mboi in the Pascal setting, where it ranks second-best. Concerning \fgari, SPOT achieves the best results in COCO and second-best in MOVi-C and MOVi-E (within the standard deviation). However, as discussed in \autoref{sec:unreliable_fgari}, \fgari is unreliable. For instance, in the Pascal dataset, the 1-block mask (\ie, the trivial solution where the entire image is covered by a single mask) achieves the highest \fgari score, doubling the score of the second-best.

\subsection{Ablations in MOVi-C}
\label{sec:movic_ablation}

In \autoref{tab:ablation_movic}, we analyze the effects of self-training and sequence permutations on the MOVi-C dataset. Both of these approaches 
enhance performance, resulting, for instance, in a 2-point increase in \mboi.

\subsection{Image encoder training stability  analysis}
\label{sec:finetuning_further_analysis}

In~\autoref{tab:self_training_further_analysis}, we notice that fine-tuning the image encoder, without self-training, can avoid training collapse by using trainable initial slots and bi-level optimization (BO-QSA~\cite{jia2022improving}). However, the achieved \mboi score, 30.7, is notably lower compared to not fine-tuning the image encoder and using randomly initialized slots, where the \mboi is 32.7 (refer to ~\autoref{tab:compSPOT} entry (b) in the main paper).
Furthermore, fine-tuning the image encoder solely with BO-QSA and without self-training exhibits a high standard deviation of 2.2, indicating there is still a training stability issue. The introduction of self-training not only enhances performance significantly (from 30.7 to 34.7) but also stabilizes the training process, as evident from the drop in standard deviation from 2.2 to 0.1. This underscores the crucial role of self-training in ensuring training stability.

\subsection{Impact of bi-level optimized query on other frameworks}
\label{sec:boqsa_other}

In \autoref{tab:boqsaexperiments}, we show that BO-QSA~\cite{jia2022improving} without our SPOT's self-training does not perform well on DINOSAUR~\cite{dinosaur} and SlotDiffusion~\cite{wu2023slotdiffusion} frameworks.

\begin{table}[!t]
\small
\centering
\setlength{\tabcolsep}{0.5pt}
\begin{tabular}{lccc|ccc|cc}
\toprule
& \mc{3}{SlotDiffusion~\cite{wu2023slotdiffusion}} & \mc{3}{DINOSAUR~\cite{dinosaur}} & \mc{2}{\cellcolor{TableColor}SPOT} \\
& Offic. & Reprod. & w/ BO & Offic. & Reprod. & w/ BO & \cellcolor{TableColor}w/o ENS & \cellcolor{TableColor}w/ ENS \\
\midrule
\mboc   & 35.0 & 36.3 & 34.7 & 38.8 & 42.1 & 39.7 & \cellcolor{TableColor}\underline{44.3} & \cellcolor{TableColor}\textbf{44.7} \\
\mboi & 31.0 & 30.5 & 29.5 & 32.3 & 32.3 & 31.5 & \cellcolor{TableColor}\underline{34.7} & \cellcolor{TableColor}\textbf{35.0}\\
\bottomrule
\end{tabular}
\caption{\emph{Results on COCO from integrating SlotDiffusion~\cite{wu2023slotdiffusion} and DINOSAUR~\cite{dinosaur} frameworks with BO-QSA, referred to as BO.} Except for \Th{DINOSAUR (Offic.)}, which is trained for 270 epochs, all others are trained for 100 epochs.}
\label{tab:boqsaexperiments}
\end{table}

\input{tex/table_pretrained_features}

\subsection{Comparing with other pre-trained image features}
\label{sec:other_encoders}

In our experiments, we used DINO~\cite{dino} features for the image encoder and reconstruction targets $Y$. We also explored MOCO-v3~\cite{chen2021empirical} and MAE~\cite{he2022masked} as alternatives, comparing SPOT with DINOSAUR on the COCO dataset. Results in~\autoref{tab:pretrained_encoder} show SPOT outperforming DINOSAUR across all examined pre-trained encoders, except with MOCO-v3 and \fgari metric. DINO excels in \mboi, \mboc, and \miou metrics, while MAE performs best in the \fgari metric.

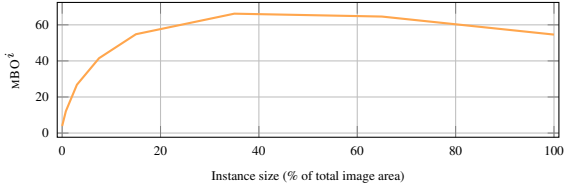
\begin{figure}[t]
\input{tex/fig_instances_size_results}
\vspace{-15pt}
\caption{\emph{Analysis of SPOT performance across different instance sizes.} We demonstrate SPOT's performance, measured in \mboi on COCO, across varied instance sizes. The instance sizes are expressed as a percentage of the total image area, categorized into distinct bins:  0-0.5\%, 0.5-1\%, 1-5\%, 5-10\%, 10-20\%, 20-50\%, 50-80\%, 80-100\%.}
\vspace{-10pt}
\label{fig:instances_size_results}
\end{figure}

\begin{figure}[tb]
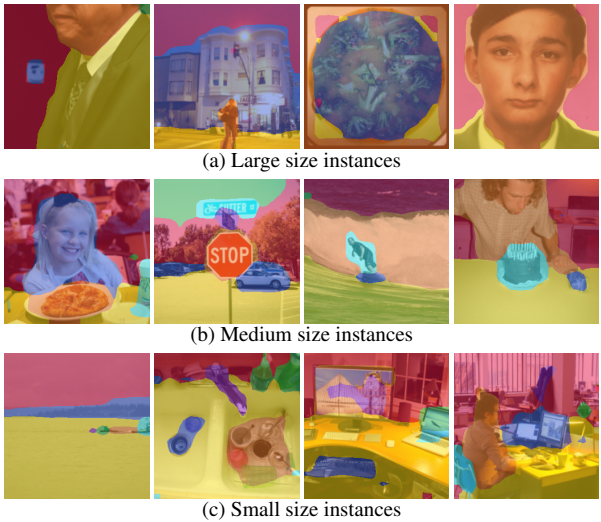

    \centering
    \include{tex/fig_large_medium_small_instances}
    \vspace{-15pt}
    \caption{Example SPOT's results on COCO 2017, using 7 slots, for large, medium and small instance sizes.
    }
    \label{fig:spot_instance_size}
    \vspace{-10pt}
\end{figure}

\subsection{Performance across different instance sizes}
\label{sec:different_sizes}

In~\autoref{fig:instances_size_results}, we present a detailed analysis of SPOT's performance across various instance sizes. We observe that SPOT performs optimally when instances occupy between 20\% and 80\% of the input image area. For larger instances (exceeding 80\%), SPOT continues to yield favorable results. Conversely, a decline in performance is noted as the size of the instances decreases. We note that this decrease in performance for smaller instance sizes is expected due to the coarse resolution of ViT encoders.

In~\autoref{fig:spot_instance_size}, we provide examples for large, medium and small instance mask sizes. For the large and medium-sized examples, we observe good performance. For small instances, there is often a tendency to be grouped together or be part of the background.

\section{Further Discussion}

\subsection{About the autoregressive decoder}
\label{sec:discuss_ar_design}

\parag{Why use an autoregressive decoder design?}
For object-centric learning, autoregressive (AR) design was shown to be superior for handling complex scenes~\cite{dinosaur,singh2022illiterate,singh2022simple} compared to spatial-broadcast MLPs that make a strong assumption that patches are independent when conditioned on slots.
In contrast, AR imposes no assumption: from the chain rule of probability, any joint distribution over random variables (e.g., patches) can be expressed as the product of conditional distributions AR-style. While not all AR factorization orderings are effective, it is crucial to note that effective ones can be defined for images. This is supported by a significant body of work in image synthesis~\cite{van2016pixel,van2016conditional,razavi2019generating,ramesh2021zero,esser2021taming,yu2022scaling,yu2023scaling,gafni2022make} and recent successes in self-supervised pre-training~\cite{el2024scalable} and generalist image models~\cite{bai2023sequential}.
Also, our AR decoder has access to the slots that encode information from the \emph{entire image}. 

\parag{Does using permutations suggest causal AR-parsing is irrelevant for 2D data?}
The use of permutations allows the AR decoder to learn from multiple factorization orders (which has shown positive effects even in text~\cite{xlnet}).
\emph{Sequence permutations aim to tackle an overfitting issue}: AR transformers rely less on input slots when predicting later tokens in the sequence. 
With the permutations, due to shared parameters across all sequence orders, the model learns to equally rely on slots across all token positions.
\emph{We perceive this permutation strategy as more natural for implementation with 2D data}, as all used orderings are equally meaningful.

\subsection{Limitations and future work}
\label{sec:limitations}

As in prior works, SPOT extracts a fixed number of slots, a constraint that should be addressed in future works. 
Additionally, integrating online teacher-student training (e.g., via momentum teachers) during the second training stage of SPOT could potentially enhance both training efficiency and performance. 
Furthermore, as depicted in~\autoref{fig:instances_size_results},
there is a decrease in performance with small-sized objects, indicating a need to explore strategies for handling higher-resolution images in future studies.
Finally, SPOT shows a preference for semantic over instance segmentation, as evidenced by its performance on the \mboi and \mboc metrics, which is linked to the broader issue of object definition ambiguity in real-world images.

\section{Implementation details}
\label{sec:more_details}
\subsection{MLP decoder}

In the MLP-based decoder, we employ the spatial broadcast mechanism as described in prior works~\cite{dinosaur,spatialbroadcast}, in which each slot is expanded into $n$ tokens that correspond to the patches. Next, learnable positional information 
is added to these tokens for spatial identification. These tokens are then processed individually by a four-layer MLP with ReLU activations. This process yields both the reconstruction and an alpha map for each slot ($d_y$ + 1 dimensions). The alpha map serves as an attention map that indicates the active regions of the slot. The final reconstructed output is obtained by aggregating these individual slot reconstructions, using the alpha maps as weighting factors.
In the MLP decoder, the slot-attention masks are the predicted alpha maps.

\subsection{SPOT models}

We employ the Adam optimizer~\citep{kingma2015adam} with $\beta_1 = 0.9$, $\beta_2 = 0.999$, no weight decay, and a batch size of 64. 
The learning rate ($lr$) follows a linear warm-up from 0 to a peak value for $10000$ training iterations and then decreases via a cosine annealing schedule.
For experiments on COCO and PASCAL using DINO and MoCo-v3 encoders and for both training stages, the peak learning rate is $4 \times 10^{-4}$ and the low value is $4 \times 10^{-7}$. For both stages with MAE on COCO and the first stage of MOVi-C/E experiments, the peak value is $2 \times 10^{-4}$ and the low value is $4 \times 10^{-5}$. For the second stage of MOVi-C/E experiments, the peak value is $2 \times 10^{-4}$ and the low value is $1.5 \times 10^{-4}$. 
For each training stage on COCO and PASCAL, we use 50 and 560 training epochs, respectively. For MOVi-C/E experiments, we use 65 and 30 epochs for the first and second stages, respectively. 

We implement our SPOT models using ViT-B/16~\cite{dosovitskiy2020image} for the encoder (by default initialized with DINO~\cite{dino}), without applying the final layer norm. In the autoregressive transformer decoder, we use 4 transformer blocks each one with 6 heads. For the MLP-based decoder, we use 2048 hidden layer size. For all experiments, we use 3 iterations in the slot attention module with the dimension of slot $d_u$ being 256 and the slot attention's MLP hidden dimension being 1024.
Unless stated otherwise, loss weight $\lambda$ is 0.005. For MLP-based decoder, MoCo-v3~\cite{chen2021empirical}, and MAE~\cite{he2022masked} encoders, $\lambda$ is 0.001. 
At the self-training stage, we fine-tune the last four ViT encoder's blocks.

Following~\cite{dinosaur}, on COCO, PASCAL, MOVi-C, and MOVi-E we use 7, 6, 11, and 24 slots, respectively. 

In the main paper, we explored a training approach inspired by CapPa~\cite{tschannen2023image}. 
To tune this approach, we experimented with different percentages of parallel decoding —25\%, 50\%, and 100\%— and observed \mboi scores of 27.8, 25.9, and 23.6, respectively. We found that a lower percentage of parallel decoding correlates with better performance. Consequently, we used a 25\% parallel decoding for the experiment in the main paper.

\subsection{Datasets and evaluation}
Here, we provide details about the employed datasets and the evaluation protocol.

\parag{COCO}
We use COCO 2017 dataset for training, which consists of 118,287 images, and evaluate SPOT on the validation set of 5,000 images. During training, we resize the minor axis of the images to 224 pixels, perform center cropping at 224$\times$224 and then random horizontal flipping with 0.5 probability. For the evaluation, we use both types of masks: instance masks for the \mboi, \miou, \fgari and segmentation masks for the \mboc metrics. 
For consistency with previous studies~\cite{dinosaur, Ji_2019_ICCV, hamilton2022unsupervised}, we scale the minor axis to 320 pixels, perform center cropping, and evaluate the masks at 320$\times$320 resolution.

\parag{PASCAL}
We use PASCAL VOC 2012 “trainaug” set for training, which consists of 10,582 images. The "trainaug" variant contains 1,464 images from the segmentation train set and 9,118 from the SBD dataset~\cite{sbd_dataset}.
We use this split, for consistency with previous works~\cite{dinosaur, Van_Gansbeke_2021_ICCV,melaskyriazi2022deep}. During training, we resize the minor axis of the images to 224 pixels, perform simple random cropping at 224$\times$224 and then random horizontal flipping with 0.5 probability.  We evaluate on the official segmentation validation set, which consists of 1,449 images. We ignore the unlabeled pixels during evaluation. Similar to COCO, we resize the minor axis to 320 pixels, perform center cropping, and evaluate the masks at 320$\times$320 resolution.

\parag{MOVi-C/E}
We transform MOVi-C/E datasets, originally consisting of videos, into image datasets by selecting nine random frames from each clip of the train set. From this process, we obtain 87,633 images for training with MOVi-C and 87,741 images with MOVi-E. During training, we resize the images at 224$\times$224 resolution. For evaluation, we use every frame from the 250 clips of the validation set consisting of 6,000 images, being consistent with prior works~\cite{dinosaur,kipf2022conditional,elsayed2022savi}. We evaluate the masks at the full 128$\times$128 resolution.

\parag{n-block Mask}: 
As in~\cite{dinosaur}, to generate block mask patterns, we first divide the image into columns and then subdivide these columns to obtain the required number of slot masks. Specifically, for MOVi-C, we utilize 3 columns to produce a total of 11 block masks. For MOVi-E, we use 4 columns to accommodate 24 masks. In the case of PASCAL VOC 2012, we use 2 columns for 6 masks. For COCO, we use 2 columns for 7 masks. The number of block masks is aligned with the number of slots utilized for each respective dataset. For the 1-block mask, we employ just one column and one mask; in other words, the entire image is covered by a single mask.

%% file: tex/fig_coco_mlp.tex
{
\footnotesize
\centering
\newcommand{\myg}[1]{\includegraphics[width=0.11\textwidth,valign=c]{#1}}
\setlength{\tabcolsep}{1pt}
\begin{tabular}{@{}cccc@{}}
   
    \myg{fig/coco/gt/000000460927_gt_matched_nobg.png} & 
    \myg{fig/coco/gt/000000553990_gt_matched_nobg.png} & 
    \myg{fig/coco/gt/000000562561_gt_matched_nobg.png} &
    \myg{fig/coco/gt/000000559956_gt_matched_nobg.png} \\
    \mc{4}{(a) Ground Truth}\\

    \mc{4}{\vspace{-2.1ex}}\\
    \myg{fig/coco/spot/000000460927_pred.png} & 
    \myg{fig/coco/spot/000000553990_pred.png} & 
    \myg{fig/coco/spot/000000562561_pred.png} &
    \myg{fig/coco/spot/000000559956_pred.png} \\

    \mc{4}{(b) SPOT
    }\\

    \mc{4}{\vspace{-2.1ex}}\\

    \myg{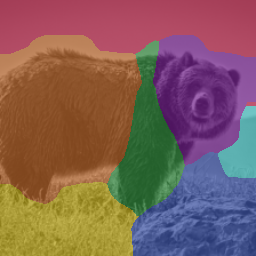} & 
    \myg{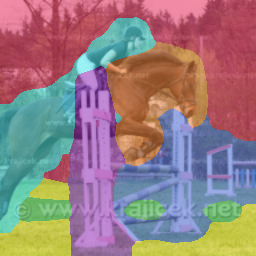} & 
    \myg{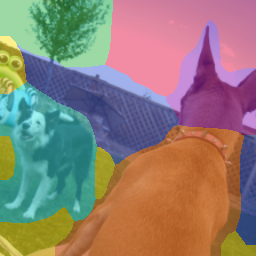} &
    \myg{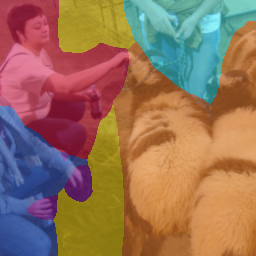} \\

    \mc{4}{(c) SPOT w/ MLP-based decoder
    }\\ 

\end{tabular}
}

%% file: tex/table_comparison_prior_work.tex
\begin{table}[t]
\small
\centering
\setlength{\tabcolsep}{1pt}
\begin{tabular}{lcccc} \toprule
\Th{Method} & \mboi & \mboc & \miou & \fgari \\ \midrule
\mc{5}{\vspace{-13pt}}\\
\mc{5}{\Th{MOVi-C}}\\ 
\mc{5}{\vspace{-14pt}}\\
\midrule

11-block Mask & 19.5\hspace{11.5pt} & - & 18.2\hspace{11.5pt} & 42.7\hspace{11.5pt} \\ 

SA~\cite{slot}$^{\dagger}$ & 26.2{\tiny $\pm$1.0} & - & - & 43.8{\tiny $\pm$0.3} \\ 

SLASH~\cite{kim2023shepherding} & - & - & 27.7{\tiny $\pm$5.9} & 51.9{\tiny $\pm$4.0} \\ 

SLATE~\cite{singh2022illiterate}$^{\dagger}$ & 39.4{\tiny $\pm$0.8} & - & 37.8{\tiny $\pm$0.7}  & 49.5{\tiny $\pm$1.4} \\ 

DINOSAUR (MLP)~\cite{dinosaur} & 39.1{\tiny $\pm$0.2} &  - & - & \bf{68.6{\tiny $\pm$0.4}} \\ 

DINOSAUR~\cite{dinosaur} & 42.4\hspace{11.5pt} & - & - & 55.7\hspace{11.5pt} \\ 

LSD~\cite{jiang2023object} & 45.6{\tiny $\pm$0.8} & -  & 44.2{\tiny $\pm$0.9} & 52.0{\tiny $\pm$3.5} \\ 

\rowcolor{TableColor}{SPOT w/o ENS (ours)} & \underline{47.0{\tiny $\pm$1.2}} & - & \underline{46.4{\tiny $\pm$1.2}} & 52.1{\tiny $\pm$3.3}/\underline{57.9{\tiny $\pm$2.0}} \\ 

\rowcolor{TableColor}{SPOT w/ ENS (ours)} & \bf{47.3{\tiny $\pm$1.2}} & - & \bf{46.7{\tiny $\pm$1.3}} & 52.3{\tiny $\pm$3.3}/\underline{57.9{\tiny $\pm$2.0}} \\\midrule

\mc{5}{\vspace{-13pt}}\\
\mc{5}{\Th{MOVi-E}}\\ 
\mc{5}{\vspace{-14pt}}\\
\midrule

24-block Mask & 20.4\hspace{11.5pt} & - & 18.8\hspace{11.5pt} & 41.9\hspace{11.5pt} \\  

SA~\cite{slot}$^{\dagger}$ & 24.0{\tiny $\pm$1.2} & - & - & 45.0{\tiny $\pm$1.7}  \\ 

SLATE~\cite{singh2022illiterate}$^{\dagger}$ & 30.2{\tiny $\pm$1.7} & - & 28.6{\tiny $\pm$1.7} & 46.1{\tiny $\pm$3.3} \\ 

DINOSAUR (MLP)~\cite{dinosaur} & 35.5{\tiny $\pm$0.2} &  - & - & \bf{65.1{\tiny $\pm$1.2}} \\ 

SlotDiffusion~\cite{wu2023slotdiffusion}   & 30.2\hspace{11.5pt} & - & 30.2\hspace{11.5pt} & \underline{60.0}\hspace{11.5pt} \\ 

LSD~\cite{jiang2023object} & 39.0{\tiny $\pm$0.5} & -  & 37.6{\tiny $\pm$0.5} & 52.2{\tiny $\pm$0.9} \\ 

\rowcolor{TableColor}{SPOT w/o ENS (ours)} & \underline{39.9{\tiny $\pm$1.1}} & - & \underline{39.0{\tiny $\pm$1.1}} & 56.4{\tiny $\pm$4.1}/\underline{59.9{\tiny $\pm$0.4}}  \\

\rowcolor{TableColor}{SPOT w/ ENS (ours)} & \bf{40.1{\tiny $\pm$1.2}} & - & \bf{39.3{\tiny $\pm$1.2}} & 56.8{\tiny $\pm$4.3}/\underline{59.9{\tiny $\pm$0.4}}   \\\midrule

\mc{5}{\vspace{-13pt}}\\
\mc{5}{\Th{PASCAL}}\\ 
\mc{5}{\vspace{-14pt}}\\
\midrule

1-block Mask & 19.1\hspace{11.5pt} & 23.0\hspace{11.5pt} & 17.4\hspace{11.5pt} & \bf{49.9}\hspace{11.5pt}  \\ 

6-block Mask & 24.7\hspace{11.5pt} & 25.9\hspace{11.5pt} & 24.2\hspace{11.5pt} & 13.9\hspace{11.5pt}  \\ 

SA~\cite{slot}$^{\dagger}$  &24.6\hspace{11.5pt} & 24.9\hspace{11.5pt} & - & 12.3\hspace{11.5pt}  \\ 

SLATE~\cite{singh2022illiterate}$^{\dagger}$  & 35.9\hspace{11.5pt} & 41.5\hspace{11.5pt} & - & 15.6\hspace{11.5pt} \\ 

CAE~\cite{lowe2022complexvalued}$^{\dagger}$ & 32.9{\tiny $\pm$0.9} & 37.4{\tiny $\pm$1.0} & - & - \\ 

DINOSAUR (MLP)~\cite{dinosaur} & 39.5{\tiny $\pm$0.1}& 40.9{\tiny $\pm$0.1} & - & 24.6{\tiny $\pm$0.2} \\ 

DINOSAUR~\cite{dinosaur}  & 44.0{\tiny $\pm$1.9} & 51.2{\tiny $\pm$1.9} & - & \underline{24.8{\tiny $\pm$2.2}} \\ 

Rotating Features~\cite{lowe2023rotating} &  40.7{\tiny $\pm$0.1} & 46.0{\tiny $\pm$0.1} & - & - \\ 

SlotDiffusion~\cite{wu2023slotdiffusion}  & \bf{50.4}\hspace{11.5pt} & \underline{55.3}\hspace{11.5pt} & - & 17.8\hspace{11.5pt}  \\ 

\rowcolor{TableColor}{SPOT w/o ENS (ours)} &
 48.1{\tiny $\pm$0.4} & \underline{{55.3{\tiny $\pm$0.4}}} & \underline{46.5{\tiny $\pm$0.4}} & 19.4{\tiny $\pm$0.7}/19.7{\tiny $\pm$0.4} \\

\rowcolor{TableColor}{SPOT w/ ENS (ours)} & \underline{{48.3{\tiny $\pm$0.4}}} & \bf{55.6{\tiny $\pm$0.4}} & \bf{46.8{\tiny $\pm$0.4}} & 19.9{\tiny $\pm$0.9}/19.7{\tiny $\pm$0.4} \\\midrule

\mc{5}{\vspace{-13pt}}\\
\mc{5}{\Th{COCO}}\\ 
\mc{5}{\vspace{-14pt}}\\
\midrule

7-block Mask  & 16.8\hspace{11.5pt} & 19.5\hspace{11.5pt} & 
15.9\hspace{11.5pt} & 22.7\hspace{11.5pt}  \\ 

SA~\cite{slot}$^{\dagger}$  & 17.2\hspace{11.5pt} & 19.2\hspace{11.5pt} & - & 21.4\hspace{11.5pt}  \\ 

SLATE~\cite{singh2022illiterate}$^{\dagger}$  & 29.1\hspace{11.5pt} & 33.6\hspace{11.5pt} & - & 32.5\hspace{11.5pt}  \\ 

DINOSAUR~\cite{dinosaur}  & 32.3{\tiny $\pm$0.4} & 38.8{\tiny $\pm$0.4} & - & 34.3{\tiny $\pm$0.5} \\ 

SlotDiffusion~\cite{wu2023slotdiffusion}   & 31.0\hspace{11.5pt} & 35.0\hspace{11.5pt} & - & \underline{37.2}\hspace{11.5pt} \\ 

Stable-LSD~\cite{jiang2023object}  & 30.4\hspace{11.5pt} & - & - & 35.0\hspace{11.5pt}  \\ 

\rowcolor{TableColor}{SPOT w/o ENS (ours)}  & \underline{34.7{\tiny $\pm$0.1}} & \underline{44.3{\tiny $\pm$0.3}} & \underline{32.7{\tiny $\pm$0.1}} & 36.6{\tiny $\pm$0.3}/\bf{37.8{\tiny $\pm$0.5}} \\

\rowcolor{TableColor}{SPOT w/ ENS (ours)} & \bf{35.0{\tiny $\pm$0.1}} & \bf{44.7{\tiny $\pm$0.3}} & \bf{33.0{\tiny $\pm$0.1}} & 37.0{\tiny $\pm$0.2}/\bf{37.8{\tiny $\pm$0.5}} \\
\bottomrule
\end{tabular}
\caption{\emph{Comparison with object-centric methods on COCO, PASCAL, MOVi-C and MOVi-E datasets}.
SPOT results are the mean and std over 3 seeds. For SPOT \fgari, we report results from both \emph{decoder/slot encoder} masks. 
DINOSAUR uses an autoregressive decoder and DINO~\cite{dino} ViT encoder (ViT-B/16 for PASCAL and MOVi-C, ViT-S/8 for COCO).
DINOSAUR-MLP uses an MLP decoder and DINO ViT encoder (ViT-B/16 for COCO and PASCAL, ViT-S/8 for MOVi-C/E).
$^{\dagger}$: COCO and PASCAL results of SA and SLATE are from \cite{wu2023slotdiffusion}, MOVi-C/E results are from \cite{dinosaur} for SA and from \cite{jiang2023object} for SLATE, PASCAL results of CAE are from \cite{lowe2023rotating}. We \textbf{bold} the best and \underline{underline} the second-best results.}
\label{tab:further_benchmark}
\end{table}

%% file: tex/table_movic_ablations.tex
\begin{table}[t]
\small
\centering
\setlength{\tabcolsep}{1pt}
\begin{tabular}{lccccc|ccc} \toprule
& \mr{2}{\Th{SP}} & \mr{2}{\Th{ST}}  & \mc{3}{\decoder} & \mc{3}{\encoder} \\ \cmidrule{4-9}

& & & \mboi & \fgiou & \fgari & \mboi & \fgiou & \fgari\\ \midrule
 
(a) & & & 45.3{\tiny $\pm$1.8} & 44.6{\tiny $\pm$1.7} & 50.6{\tiny $\pm$4.3} & 42.8{\tiny $\pm$0.1} & 42.0{\tiny $\pm$0.1}  & 55.4{\tiny $\pm$0.6} \\ 
 
(b) & \cmark &  & 46.1{\tiny $\pm$0.9} &  45.2{\tiny $\pm$0.9} & 51.8{\tiny $\pm$2.7} & 42.5{\tiny $\pm$0.1}  & 41.6{\tiny $\pm$0.1}  & 57.6{\tiny $\pm$0.7}\\

\rowcolor{TableColor}(c) & \cmark & \cmark & \bf{47.3{\tiny $\pm$1.2}}
& \bf{46.7{\tiny $\pm$1.3}} & \bf{52.3{\tiny $\pm$3.3}} & \bf{46.2{\tiny $\pm$0.8}} & \bf{45.4{\tiny $\pm$0.7}} & \bf{57.9{\tiny $\pm$2.0}} \\
\bottomrule
\end{tabular}
\vspace{-8pt}
\caption{\emph{Ablation study on MOVi-C}. Metrics for slot masks generated by \decoder and \encoder. Results are mean and standard dev. over 3 seeds. \Th{SP}: sequence permutation with ensembling of nine permutations at test-time, \Th{ST}: self-training.}
\label{tab:ablation_movic}
\vspace{-5.8pt}
\end{table}

%% file: tex/table_self_training_further_analysis.tex
\begin{table}[!t]\centering
\small
\centering
\centering
\begin{tabular}{ccc} \toprule
 \mr{1}{\Th{BO-QSA~\cite{jia2022improving}}} & \mr{1}{\Th{ST}} & \mr{1}{\mboi} \\ 
 \midrule
  & & \Th{Collapse} \\
 \cmark & & 30.7{\tiny $\pm$2.2} \\
\rowcolor{TableColor} \cmark & \cmark & \bf{34.7{\tiny $\pm$0.1}}\\ \bottomrule
\end{tabular}
\vspace{-8pt}
\caption{\emph{Image encoder training stability results on COCO}. 
Results are the mean and standard deviation of the decoder's \mboi over 3 seeds.
\Th{ST}: self-training studies the impact of the distillation loss $L_{\mathrm{ATT}}$,  and \Th{BO-QSA} the impact of using trainable initialization of slots along with bi-level optimization~\cite{jia2022improving, chang2022object}.
All models use sequence permutation in the autoregressive decoder.
}
\label{tab:self_training_further_analysis}
\vspace{-5.8pt}
\end{table}

%% file: tex/table_pretrained_features.tex
\begin{table}[t]
\small
\centering
\setlength{\tabcolsep}{2.1pt}
\begin{tabular}{l|l|cccc} \toprule
 \Th{Encoder}& \Th{Method} & \mboi & \mboc & \miou & \fgari\\ \midrule
 \mr{3}{DINO} & DINOSAUR & 31.6{\tiny $\pm$0.7} & 39.7{\tiny $\pm$0.9} & - & 34.1{\tiny $\pm$1.0}  \\ 
 & \cellcolor{TableColor}SPOT w/o ENS & \cellcolor{TableColor}34.7{\tiny $\pm$0.1} & \cellcolor{TableColor}44.3{\tiny $\pm$0.3} & \cellcolor{TableColor}32.7{\tiny $\pm$0.1} & \cellcolor{TableColor}36.6{\tiny $\pm$0.3}  \\
& \cellcolor{TableColor}SPOT w/ ENS & \cellcolor{TableColor}\bf{35.0{\tiny $\pm$0.1}} & \cellcolor{TableColor}\bf{44.7{\tiny $\pm$0.3}} & \cellcolor{TableColor}\bf{33.0{\tiny $\pm$0.1}} & \cellcolor{TableColor}37.0{\tiny $\pm$0.2} \\\midrule

 \mr{3}{MoCo-v3} & DINOSAUR & 31.4{\tiny $\pm$0.2} & 38.5{\tiny $\pm$0.5} & - & 35.2{\tiny $\pm$0.2} \\ 
 & \cellcolor{TableColor}SPOT w/o ENS & \cellcolor{TableColor}32.7{\tiny $\pm$0.2} &	\cellcolor{TableColor}41.7{\tiny $\pm$0.4} & \cellcolor{TableColor}30.7{\tiny $\pm$0.2}	& \cellcolor{TableColor}34.4{\tiny $\pm$0.2}  \\
& \cellcolor{TableColor}SPOT w/ ENS & \cellcolor{TableColor}32.9{\tiny $\pm$0.2} &\cellcolor{TableColor}42.0{\tiny $\pm$0.4} &	\cellcolor{TableColor}30.9{\tiny $\pm$0.2} & \cellcolor{TableColor}34.8{\tiny $\pm$0.3}  \\\midrule

 \mr{3}{MAE} & DINOSAUR & 30.2{\tiny $\pm$1.8} & 33.2{\tiny $\pm$1.8} & - & 32.8{\tiny $\pm$3.7} \\ 
 & \cellcolor{TableColor}SPOT w/o ENS & \cellcolor{TableColor}33.3{\tiny $\pm$0.3} & \cellcolor{TableColor}40.7{\tiny $\pm$0.7} & \cellcolor{TableColor}31.4{\tiny $\pm$0.3} & \cellcolor{TableColor}37.5{\tiny $\pm$1.0} \\
& \cellcolor{TableColor}SPOT w/ ENS & \cellcolor{TableColor}33.4{\tiny $\pm$0.3} & \cellcolor{TableColor}40.9{\tiny $\pm$0.7} & \cellcolor{TableColor}31.6{\tiny $\pm$0.3} & \cellcolor{TableColor}\bf{37.7{\tiny $\pm$1.0}} \\
\bottomrule
\end{tabular}
\vspace{-8pt}
\caption{\emph{Evaluation with various pre-trained encoders on COCO}.
SPOT results are the mean and standard deviation over 3 seeds.
}
\label{tab:pretrained_encoder}
\vspace{-10pt}
\end{table}

%% file: tex/fig_instances_size_results.tex
\begin{minipage}[t]{0.45\linewidth}

\small
\centering
\pgfplotstableread{
	percent mboi
        0  3.9
        0.75  12.1
	3     26.8
	7.5   41.5
	15    54.8
	35    66.2
	65    64.6
	100    54.6

}{\epoch}
\extfig{epoch}{
\begin{tikzpicture}[%
	lab/.style={fill=white,fill opacity=.8},
]
\begin{axis}[%
	width=2.2\linewidth,
	height=0.9\linewidth,
	font=\tiny,
	xlabel={Instance size (\% of total image area)},
	ylabel={\mboi},
        xmin=-1,   
        xmax=101, 
]
	\addplot[orange!70] table[x=percent,y=mboi] \epoch;

\end{axis}
\end{tikzpicture}
}

\end{minipage}

%% file: tex/fig_large_medium_small_instances.tex
{
\footnotesize
\centering
\newcommand{\myg}[1]{\includegraphics[width=0.11\textwidth,valign=c]{#1}}
\setlength{\tabcolsep}{1pt}
\begin{tabular}{@{}cccc@{}}
   
    \myg{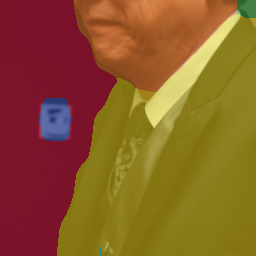} & 
    \myg{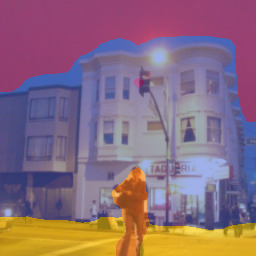} & 
    \myg{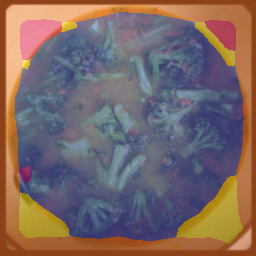} &
    \myg{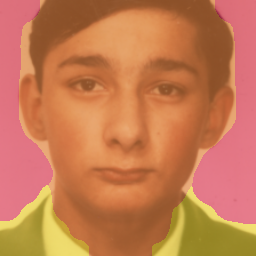} \\
    \mc{4}{(a) Large size instances}\\

    \mc{4}{\vspace{-2.1ex}}\\
    \myg{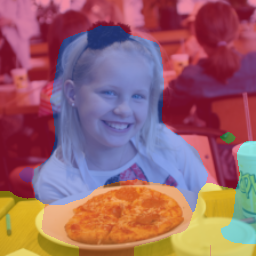} & 
    \myg{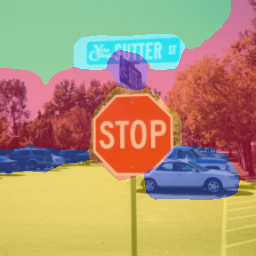} & 
    \myg{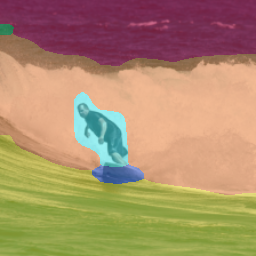} &
    \myg{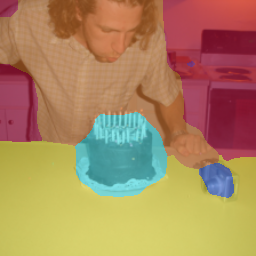} \\

    \mc{4}{(b) Medium size instances
    }\\

    \mc{4}{\vspace{-2.1ex}}\\

    \myg{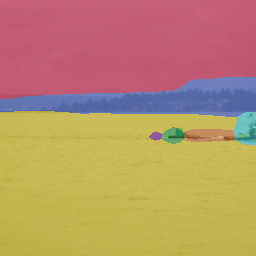} & 
    \myg{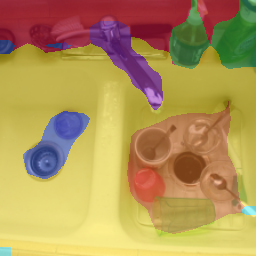} & 
    \myg{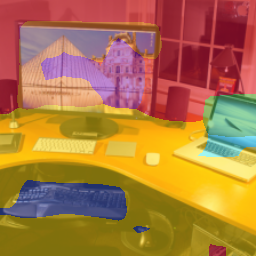} &
    \myg{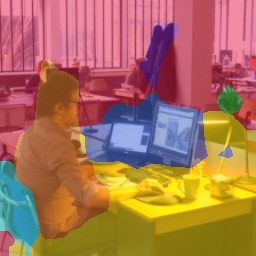} \\

    \mc{4}{(c) Small size instances
    }\\ 

\end{tabular}
}